\documentclass[10pt,twocolumn,letterpaper]{article}

\usepackage{cvpr}              %

\usepackage{array}
\usepackage[table]{xcolor}
\usepackage{graphicx}
\usepackage{amsmath}
\usepackage{amssymb}
\usepackage{booktabs}
\usepackage{float}
\usepackage{comment}
\usepackage[accsupp]{axessibility}
\usepackage{tikz}

\usepackage{algorithm}
\usepackage{algpseudocode}

\usepackage{tabularx}
\usepackage{multirow}
\usepackage{adjustbox}
\usepackage{makecell}
\usepackage{environ}
\usepackage{subcaption} %
\usepackage{placeins}
\usepackage{listings}
\usepackage[
  cachedir={mintedmain},
  frozencache
]{minted}

\newcolumntype{Y}{>{\centering\arraybackslash}X}

\newif\ifisdraft
\isdraftfalse
\isdrafttrue
\NewEnviron{drafty}{%
    \ifisdraft
        {\color{lightgray} \BODY}
    \fi
}

\NewEnviron{TODO}{%
    \ifisdraft
        {\color{red} \BODY}
    \fi
}

\newcommand{\lightmidrule}{
\arrayrulecolor{gray!50!white}
\midrule
\arrayrulecolor{black}
}

\definecolor{codeblue}{RGB}{70,136,140}     %
\definecolor{codefunc}{RGB}{220,20,120}     %
\definecolor{codesign}{RGB}{40,40,255}      %
\definecolor{algoyellow}{RGB}{255,215,0}

\lstdefinelanguage{PythonFuncColor}{
  language=Python,
  keywordstyle=\color{blue}\bfseries,
  commentstyle=\color{codeblue},
  stringstyle=\color{orange},
  showstringspaces=false,
  basicstyle=\ttfamily\small,
  literate=
    {*}{{\color{codesign}*}}{1}
    {-}{{\color{codesign}-}}{1}
    {+}{{\color{codesign}+}}{1}
    {/}{{\color{codesign}/}}{1}
    {=}{{\color{black}=}}{1}
    {dataloader}{{\color{codefunc}dataloader}}{1}
    {sample_t}{{\color{codefunc}sample\_t}}{1}
    {randn}{{\color{codefunc}randn}}{1}
    {randn_like}{{\color{codefunc}randn\_like}}{1}
    {rand_like}{{\color{codefunc}rand\_like}}{1}
    {sigmoid}{{\color{codefunc}sigmoid}}{1}
    {min}{{\color{codefunc}min}}{1}
    {abs}{{\color{codefunc}abs}}{1}
    {where}{{\color{codefunc}where}}{1}
    {lognorm}{{\color{codefunc}lognorm}}{1}
    {jvp}{{\color{codefunc}jvp}}{1}
    {stopgrad}{{\color{codefunc}stopgrad}}{1}
    {l2_loss}{{\color{codefunc}l2\_loss}}{1}
    {net_fn}{{\color{codefunc}net}}{1}
    {net}{{\color{codefunc}net}}{1}
    {return}{{\color{codefunc}return}}{1}
}
\definecolor{cvprblue}{rgb}{0.21,0.49,0.74}
\usepackage[pagebackref,breaklinks,colorlinks,allcolors=cvprblue]{hyperref}

\def\paperID{REMOVED} %
\def\confName{CVPR}
\def\confYear{2026}

\newcommand{\ours}{Patch Forcing}

\title{Denoising, Fast and Slow: Difficulty-Aware \\Adaptive Sampling for Image Generation}

\author{
    Johannes Schusterbauer\thanks{Equal Contribution.} \qquad Ming Gui\footnotemark[1] \\ Yusong Li \qquad
    Pingchuan Ma \qquad Felix Krause \qquad Bj\"orn Ommer\\[0.8em]
    CompVis @ LMU Munich, Munich Center for Machine Learning (MCML)
}

\begin{document}
\maketitle

\begin{abstract}
Diffusion- and flow-based models usually allocate compute uniformly across space, updating all patches with the same timestep and number of function evaluations. While convenient, this ignores the heterogeneity of natural images: some regions are easy to denoise, whereas others benefit from more refinement or additional context.
Motivated by this, we explore patch-level noise scales for image synthesis.
We find that naively varying timesteps across image tokens performs poorly, as it exposes the model to overly informative training states that do not occur at inference.
We therefore introduce a timestep sampler that explicitly controls the maximum patch-level information available during training, and show that moving from global to patch-level timesteps already improves image generation over standard baselines.
By further augmenting the model with a lightweight per-patch difficulty head, we enable adaptive samplers that allocate compute dynamically where it is most needed.
Combined with noise levels varying over both space and diffusion time, this yields \emph{Patch Forcing} (PF), a framework that advances easier regions earlier so they can provide context for harder ones. PF achieves superior results on class-conditional ImageNet, remains orthogonal to representation alignment and guidance methods, and scales to text-to-image synthesis. Our results suggest that patch-level denoising schedules provide a promising foundation for adaptive image generation.
Code available here: \small \url{https://github.com/CompVis/patch-forcing}.
\end{abstract}

\begin{figure}
    \centering
    \includegraphics[width=0.95\linewidth]{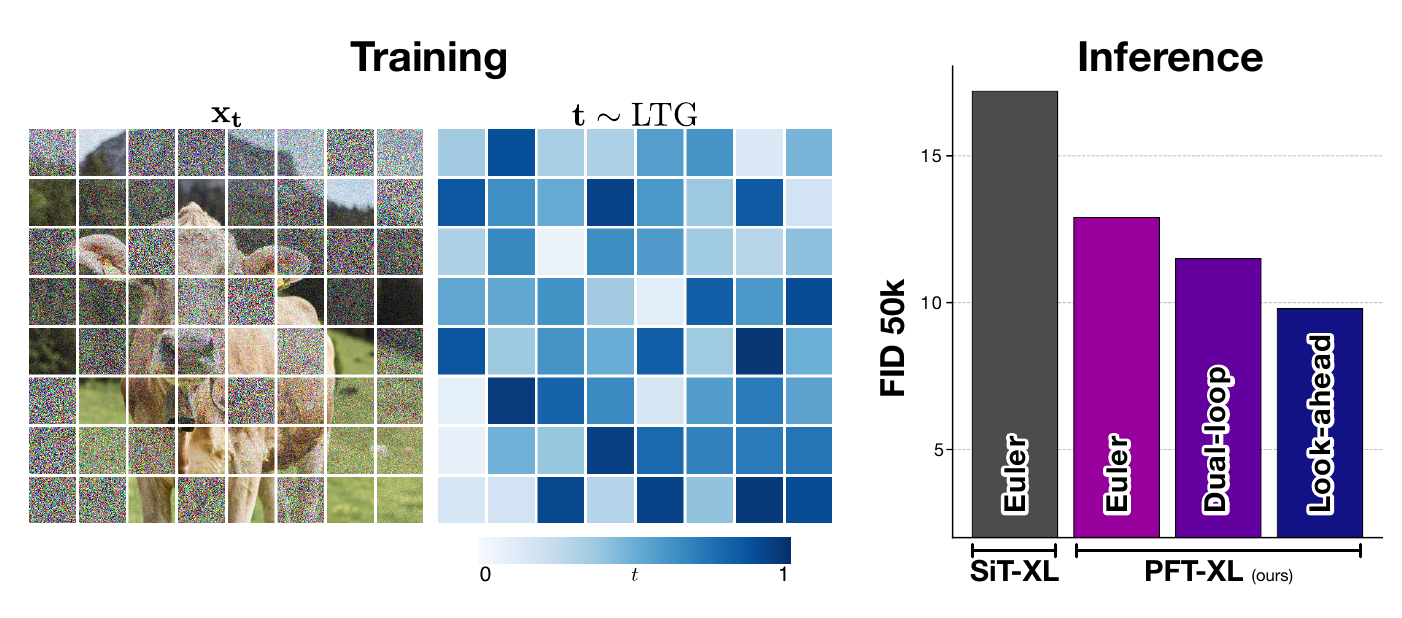}
    \vspace{-0.3cm}
    \caption{
    \textbf{Patch Forcing overview.}
    We show that heterogeneous patch timesteps during training can improve image generation, even with basic Euler sampling, but only when paired with the right timestep sampling strategy.
    This further enables adaptive inference strategies that allocate more compute to difficult patches, yielding further gains under the same sampling budget.
    }
    \vspace{-0.3cm}
    \label{fig:front-page-figure}
\end{figure}

\section{Introduction}
\label{sec:intro}

\begin{figure*}[t]
\centering
\includegraphics[width=\linewidth]{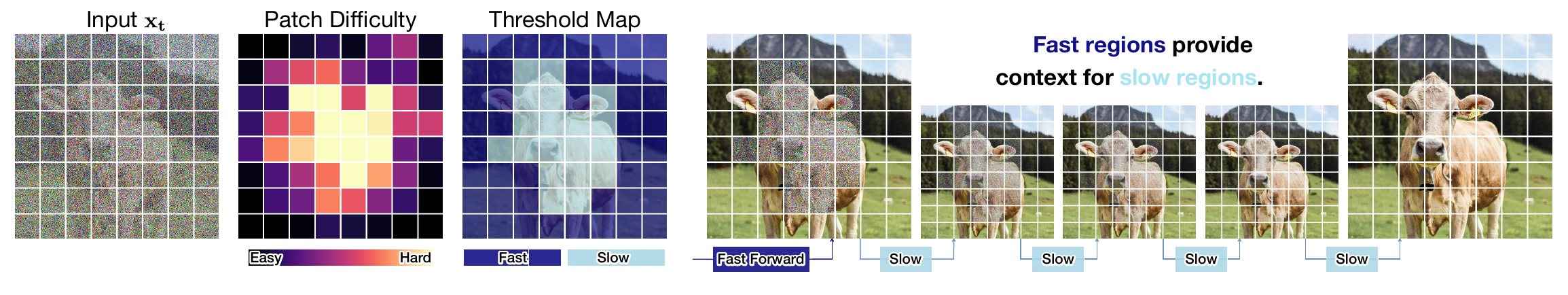}
\definecolor{easy}{HTML}{1d008b}
\definecolor{hard}{HTML}{a6d9e8}
\vspace{-0.7cm}
\caption{\textbf{Patch Forcing inference.}
Given the input, our model predicts a patch difficulty map over the velocity field, separating \textbf{\textcolor{easy}{confident}} (easy) from \textbf{\textcolor{hard}{uncertain}} (hard) denoising regions. In our difficulty-aware samplers, \textbf{\textcolor{easy}{confident regions} move faster} along the denoising trajectory and provide intermediate, cleaner \textbf{context for \textcolor{hard}{uncertain regions}}, improving generation performance.
}
\label{fig:concept}
\vspace{-0.3cm}
\end{figure*}

Most modern diffusion and flow-based image generators spend their compute uniformly: at every denoising step, every spatial location is updated with the same number of function evaluations and the same global noise level \cite{ma2024sit, peebles2023scalable-dit, rombach2022high-sd-ldm}. This design is convenient, but it implicitly assumes that \textit{all regions of an image are equally difficult} to denoise. In practice, image \textit{regions are highly heterogeneous}. Large, low-frequency backgrounds or saturated regions are easy, while thin structures, object boundaries, small text, or occlusion boundaries remain ambiguous until late in the denoising process.
Treating these regions identically forces the model to spend unnecessary compute on easy areas, while insufficiently allocating resources to regions that a) require more refinement and b) would benefit from more context. This need for additional context for hard regions or tasks is not unique to images: large language models also improve when given better contexts~\cite{brown2020language}.

In image synthesis, additional context has predominantly been introduced by means of better conditioning signals, either spatially via depth maps \cite{esser2023structure_gen1,zhang2023adding_controlnet} or globally, through representations~\cite{yurepresentation_repa, gui2025adapting, zheng2025diffusion_rae}, or (improved) text \cite{chefer2023attend}.
In contrast, in this work, we explore adapting the denoising process itself so that it produces its own contextual information to the benefit of more complex regions.
This idea relates to image editing~\cite{efros2023image,barnes2009patchmatch,meng2021sdedit,huang2025diffusion,kulikov2025flowedit}, inpainting methods~\cite{bertalmio2000image,jam2021comprehensive,lugmayr2022repaint}, and masked modeling~\cite{he2021masked,chang2022maskgit}, which leverage partially observed images to reconstruct or refine local regions. However, unlike these approaches, which rely on ground-truth information during inference, our goal is to produce contextual guidance internally, allowing certain patches to denoise more quickly and providing context to other, noisier regions.
This dynamic allocation serves two goals. On one hand, it makes better use of a given compute budget by focusing on the most helpful updates. On the other hand, it enables a novel test-time adaptation: by letting confident patches move ahead and conditioning uncertain ones on their ``future'' states, the model can actively use the context it generated to reduce ambiguity.

We implement our approach using \emph{Diffusion Forcing}~\cite{chen2024diffusion_forcing}, a method that applies different noise levels during training and sampling.
Although Diffusion Forcing was first developed for video generation, Spatial Reasoning Models (SRM)~\cite{wewer2025spatial_srm} adapt it to the image domain for spatial reasoning tasks, such as Sudoku, enabling the model to first ``solve'' easy cells and then condition on them when resolving harder ones. 
We build on SRM by introducing patch-level diffusion forcing for image synthesis, where spatial patches within an image follow distinct noise trajectories and exchange contextual information across space and time.
To determine which patches should be denoised earlier than others, we use a learned \emph{uncertainty} head that predicts patch-wise uncertainty for the model's velocity estimate, similar to~\cite{wewer2025spatial_srm, nichol2021improvedDDPM}. We explicitly repurpose this score as a measure of local \emph{patch difficulty} rather than as epistemic or aleatoric uncertainty~\cite{seitzerpitfalls}: higher values denote harder regions that require more compute and context. For brevity, we use the term ``uncertainty'' interchangeably with patch difficulty throughout the paper.

\begin{figure*}[t]
    \centering
    \includegraphics[width=\linewidth]{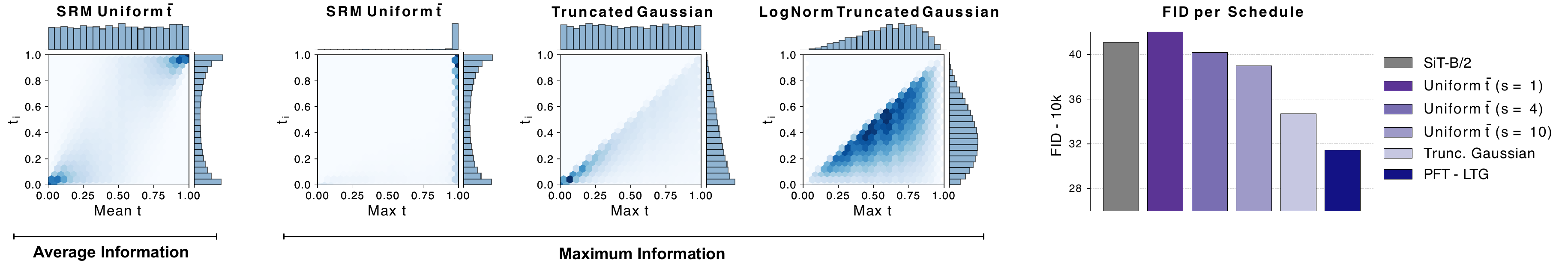}
    \vspace{-0.55cm}
    \caption{
    \textbf{Timestep sampling during training.}
    The per-sample $t_{\max}$ distribution under SRM’s~\cite{wewer2025spatial_srm} uniform $\bar{t}$ schedule places very high mass near $t=1$, meaning that for almost every training sample, there is at least one patch that is (near-)fully denoised. This introduces context leakage and a train-test mismatch. Our Logit-Normal Truncated Gaussian (LTG) sampler reduces this effect by controlling $t_{\max}$ while spreading individual $t_i$ more evenly, yielding stronger image-synthesis performance.
    }
    \label{fig:srm-comparison}
    \vspace{-0.2cm}
\end{figure*}

With patch-level noise scales and an uncertainty head, we introduce our \emph{Patch Forcing} (PF) approach, and base it on three key findings. First, \emph{more context helps generation}, as shown by reduced validation losses with increased context. Second, we show that the predicted uncertainty correlates with patch difficulty: regions with low uncertainty tend to exhibit lower validation losses than those with high uncertainty. And third, we find that providing \emph{more context reduces uncertainty}. As additional patch-level context becomes available, the uncertainty in ambiguous regions decreases.
We therefore make the following contributions.

We introduce a \textit{patch-wise timestep sampling strategy} for image generative models that addresses the train-test gap of naive patch-based Diffusion Forcing~\cite{chen2024diffusion_forcing}, where independent sampling concentrates per-sample means near $\bar{t}\approx 0.5$ (\Cref{sec:pf:training}).
This leads to training with partial context, while inference starts from pure noise.
While SRM~\cite{wewer2025spatial_srm} attempts to correct this distribution by uniformly sampling the mean and constructing per-patch timesteps through a recursive binning procedure, this added complexity both increases computational cost and still maintains context within each training sample, an effect discussed in detail in~\ref{sec:method}. Rather than controlling the \emph{average} information during training, our proposed solution controls the \emph{maximum} information. By sampling from the lower half of a Gaussian distribution, we ensure the model performs well even when no information is available. This simpler strategy better matches the inference distribution and outperforms SRM~\cite{wewer2025spatial_srm} and baselines such as SiT~\cite{ma2024sit}.

In addition, the Patch Forcing framework incorporates a learned patch difficulty head. This enables \textit{uncertainty-aware} sampling, in which compute is dynamically allocated: high-confidence patches take larger denoising steps, while high-uncertainty patches receive more refinement. We provide empirical evidence showing that predicted uncertainty maps closely match patchwise synthesis difficulty and that additional spatial context consistently reduces both predicted uncertainty and validation error in ambiguous regions. These findings explain the advantage of our adaptive samplers over normal uniform Euler sampling.

We show that our \emph{Patch Forcing Transformer} (PFT) consistently improves performance on the standard ImageNet generation benchmark~\cite{russakovsky2015imagenet}, and demonstrate that it enables sampling strategies to improve quality. Furthermore, we show that our framework and difficulty-aware samplers naturally extend to text-to-image synthesis.

\section{Related works}
\label{sec:related}

\paragraph{Diffusion and Flow Matching Models.}
Diffusion~\cite{ho2020denoising,song2020score,song2021denoising-ddim,nichol2021improvedDDPM} and flow matching models~\cite{lipman2022flow,albergo2023stochastic,liu2022rectifiedflow} form the foundation for modern generative modeling.
On the architectural front, Diffusion Transformers (DiTs)~\cite{peebles2023scalable-dit} replaced U-Net architectures in the latent diffusion model paradigm~\cite{rombach2022high-sd-ldm} with transformers operating on spatial patches. Scalable Interpolant Transformers (SiT)~\cite{albergo2023stochastic,ma2024sit} unified diffusion and flow models through the interpolant framework.
A critical challenge for these models is controlling generation to balance sample quality and diversity.
While Classifier-Free Guidance (CFG) \cite{ho2021classifier_cfg} improves generation quality, it suffers from quality-diversity trade-offs and can push samples \textit{away from the data manifold} at high guidance scales. Attention-based guidance methods, \eg, SAG~\cite{hong2023improving}, PAG~\cite{ahn2024self}, and ERG~\cite{ifriqi2025entropy}, improve upon this by modifying the attention mechanism itself.
These guidance approaches steer the denoising trajectory through contrasting predictions computed at inference.

Rather than steering generation through guidance signals, \ours{} generates context internally during the denoising process to improve generation quality: confident regions denoise faster and provide contextual information to guide uncertain regions. Importantly, this effect is different and orthogonal to guidance techniques~\cite{ho2021classifier_cfg, dhariwal2021diffusion}. To enable this adaptive strategy, we must identify which regions are confident and which require additional refinement.

\paragraph{Uncertainty in Generative Models.}
Recent generative models produce confidence scores indicating the difficulty of the generated content. 
MaskGIT~\cite{chang2022maskgit} iteratively refines discrete tokens based on prediction confidence, demonstrating that spatial heterogeneity in confidence enables adaptive generation. 
Token-Critic~\cite{lezama2022improved} extends this by training a separate critic network to predict token plausibility. 
Spatial Reasoning Models (SRM)~\cite{wewer2025spatial_srm} learn the uncertainty as standard deviation and minimize the negative log-likelihood of the ground truth noise, similar to heteroscedastic uncertainty estimation~\cite{seitzerpitfalls}.
Score-based diffusion models~\cite{song2020score} similarly perform adaptive refinement via predictor-corrector sampling, where stochastic corrector steps refine uncertain regions after each deterministic prediction, showing that heterogeneous refinement improves sample quality. 
We repurpose the variance prediction as a proxy for patch difficulty and use it as a guidance signal for patch selection, enabling adaptive step and patch budgets.

\paragraph{Adaptive Denoising.}
Acting on uncertainty predictions requires mechanisms to determine which patches denoise at which timesteps.
Diffusion Forcing~\cite{chen2024diffusion_forcing} introduces heterogeneous per-frame timesteps, allowing lower-noise frames to guide noisier ones.
AdaDiff~\cite{zhang2025adadiff} adapts computation using uncertainty, but operates globally at each step rather than spatially locally.
PatchScaler~\cite{liu2025patchscaler} groups patches by reconstruction difficulty for super-resolution.
Adaptive Non-Uniform Timestep Sampling~\cite{kim2024adaptive} improves training efficiency by prioritizing globally important timesteps, but does not consider patch-wise timestep allocation.
For inference acceleration, Region-Adaptive Sampling (RAS)~\cite{liu2025region} updates salient regions and caches stable ones based on attention, prioritizing efficiency over fidelity. 
In contrast, \ours{} is a \emph{training-time} framework with patch-wise timesteps and difficulty prediction, enabling spatially local adaptation during generation. Building on this, we further improve fidelity by advancing confident patches to provide context for harder ones. These components complement inference-time caching methods such as RAS.

\begin{figure*}[tb]
    \centering
    \includegraphics[width=0.98\linewidth]{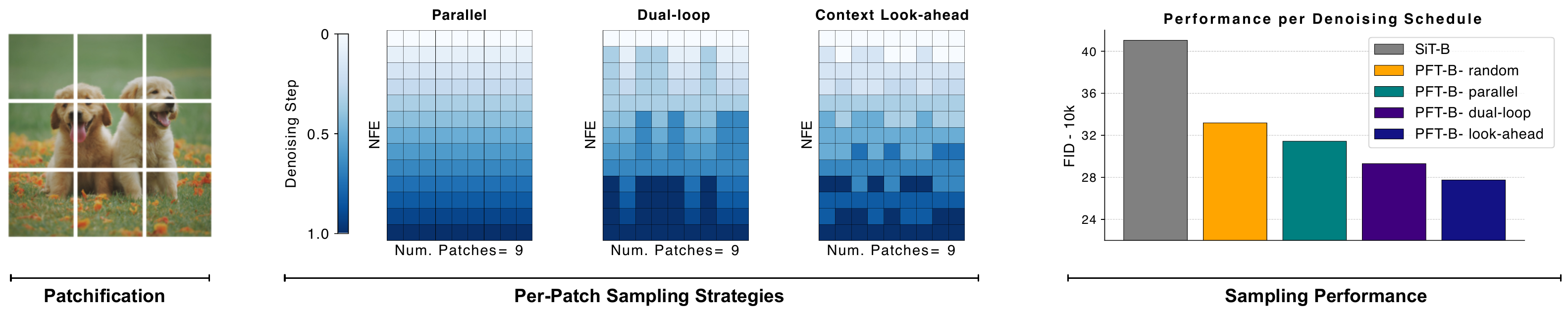}
    \vspace{-0.25cm}
    \caption{\textbf{Illustation of adaptive samplers}. \textit{Left}: $3 \times 3$ patchification of an image. \textit{Middle}: For $9$ patches, we visualize different sampling strategies. In Dual-Loop and Context Look-ahead, confident patches move faster than uncertain patches. \textit{Right}: Generation performance across scheduling strategies. While Patch Forcing already outperforms the SiT baseline, the ordering choice is important: structured approaches like dual-loop and look-ahead yield better results than randomly selecting confident patches.
    }
    \label{fig:schedule-vs-fid}
    \vspace{-0.3cm}
\end{figure*}

\section{Method}
\label{sec:method}
In this section, we present \ours{} (PF), a flexible framework for spatially adaptive image generation that explores Diffusion Forcing at the image patch level. By allowing different regions of an image to follow different denoising trajectories, PF enables confident patches to be denoised earlier and provide context for more uncertain ones. We first review the relevant background on Flow Matching, Diffusion Forcing, and Spatial Reasoning models in \cref{sec:pf:preliminaries}. We then describe PF’s training-time timestep sampling strategy and how it incorporates a learned uncertainty head to determine patch-wise denoising priorities in \cref{sec:pf:training}. Finally, in \cref{sec:pf:sampling}, we show how the predicted uncertainty can be used to construct uncertainty-aware inference strategies.

\subsection{Preliminaries}
\label{sec:pf:preliminaries}

\paragraph{Flow Matching} (FM) models \cite{liu2022rectifiedflow, albergo2023stochastic, lipman2022flow} gradually deteriorate data samples using a predefined noising schedule and then learn a neural network to reverse this process. We adopt the setting from \citet{ma2024sit}, where $\mathbf{x}_0$ represents noise and $\mathbf{x}_1$ corresponds to data.
The interpolant on the continuous timesteps $t \in [0, 1]$ is defined as
\begin{equation}
    \mathbf{x}_t = t \mathbf{x}_1 + (1-t) \mathbf{x}_0,
\end{equation}
where $\mathbf{x}_0 \sim \mathcal{N}(0, \mathbf{I})$ and $\mathbf{x}_1$ is a data sample.
We train a denoising model $\mathbf{v}_\theta$ to regress a vector field along the noising trajectory, following the linear path that points from $\mathbf{x}_0$ to $\mathbf{x}_1$ using the following objective:
\begin{equation}
    \mathcal{L}_\text{FM}=\mathbb{E}_{t,\mathbf{x}_0,\mathbf{x}_1} || (\mathbf{x}_1 - \mathbf{x}_0) - \mathbf{v}_\theta(\mathbf{x}_t, t) ||^2.
    \label{eq:fm_loss}
\end{equation}

\paragraph{Diffusion Forcing} \cite{chen2024diffusion_forcing}, originally introduced for sequential data, such as video, it assigns independent timesteps to each element along the temporal dimension. Instead of a single global timestep, each frame is conditioned on its own noise level, enabling heterogeneous denoising across time. This allows the model to leverage partially denoised information during generation.

\paragraph{Spatial Reasoning Models} (SRMs) \cite{wewer2025spatial_srm} adapts Diffusion Forcing~\cite{chen2024diffusion_forcing} for spatial reasoning tasks, showing that generation order can be learned and leveraged to improve spatial reasoning. 
They observe that independently sampling $t_i \in \mathcal{U}(0,1)$ for each patch leads to the mean $\bar{t}$ following a Bates distribution, which is heavily concentrated around $0.5$ and deviates substantially from the expected inference-time distribution.
To address this, they propose sampling $\bar{t}$ uniformly and then generating patch-specific timesteps $t_i$ from $p(t_i \mid \bar{t})$ using a recursive allocation sampling scheme.

\subsection{\ours{} Training}
\label{sec:pf:training}

\begin{figure}[tb]
    \centering

    \begin{subfigure}[t]{0.54\linewidth}
        \begin{minipage}[t]{\linewidth}
        \centering
        \includegraphics[width=\linewidth]{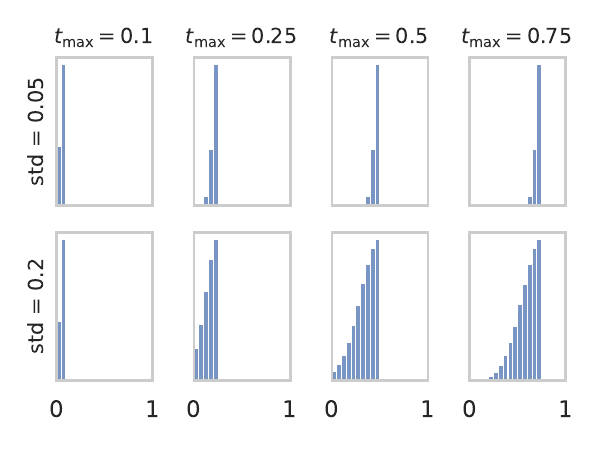}
        \caption{Patch-wise timesteps given $t_\text{max}$}
        \label{fig:truncate-gaussian-viz}
        \end{minipage}
    \end{subfigure}
    \hfill
    \begin{subfigure}[t]{0.4\linewidth}
        \begin{minipage}[t]{\linewidth}
        \centering
        \includegraphics[width=\linewidth]{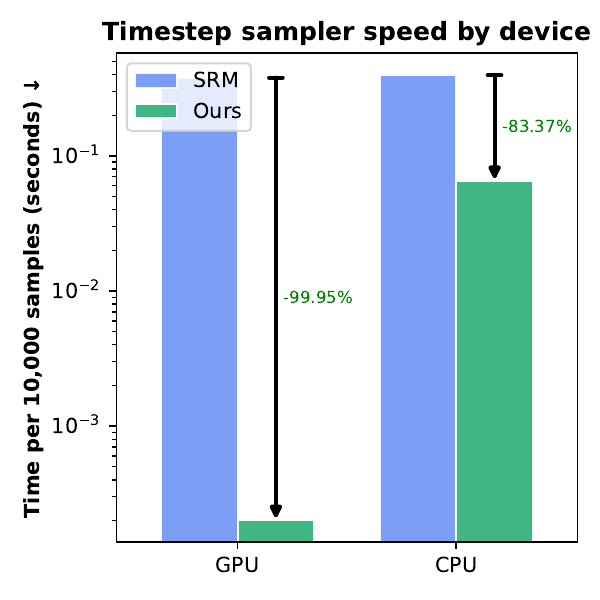}
        \caption{Uniform $\bar{t}$ sampler vs ours}
        \label{fig:our_scheduler_vs_SRM}
        \end{minipage}
    \end{subfigure}

    \caption{\textbf{Our Logit-Normal Truncate Gaussian (LTG) sampler.} \textbf{(a)} $t_{\max}$ determines where the truncated Gaussian is located, while $\text{std}$ controls its spread. Setting $\text{std}=0$ collapses the distribution to a Dirac delta distribution, reducing our method to standard Flow Matching. \textbf{(b)} Our sampler enables fast sampling via parallelization instead of SRM's recursive sampling.}
    \label{fig:side-by-side-timesteps}
    \vspace{-0.2cm}
\end{figure}

\paragraph{Noise Level Sampling}

Unlike conventional diffusion image generation models that use a uniform timestep $t \in \mathbb{R}$ across all patches, we introduce per-patch timestep conditioning for heterogeneous denoising. Concretely, given an image or latent representation $\mathbf{x} \in \mathbb{R}^{C \times H \times W}$ and a transformer patch size $p$, we sample a timestep for each patch, yielding $\mathbf{t} \in \mathbb{R}^{(H/p) \times (W/p)}$.
Since Diffusion Transformers~\cite{ma2024sit,peebles2023scalable-dit} already broadcast scalar timesteps through Adaptive LayerNorm~\cite{zhu2024sd-dit} across spatial locations, this only requires a minimal adaptation: we simply extend the same mechanism to support spatially varying timesteps.

A natural starting point is to sample $t_i \sim \mathcal{U}(0, 1)$ of $\mathbf{t}$ independently per patch.
However, as observed by \cite{wewer2025spatial_srm}, this induces a timestep distribution concentrated around $\bar{t} \approx 0.5$, such that training samples almost always contain substantial clean signal.
To mitigate this, SRM enforces a uniform distribution over the mean timestep $\bar{t}$, followed by a recursive allocation scheme to generate per-patch timesteps $t_i$.
While this successfully controls the \emph{average} information per training sample, it does not constrain the \emph{maximum} information. In practice, the most informative timestep $t_{\text{max}}$ still remains close to $t = 1$ with high probability (\cref{fig:srm-comparison}, left), meaning that during training there are almost always patches that are fully denoised. This results in overly informative training states and introduces a \emph{train-test gap}, as image generation at inference typically starts from pure noise.

\begin{figure}[t]
    \centering \footnotesize
    \setlength\tabcolsep{0.5pt}
    \newcommand{\imagepng}[2]{
        \includegraphics[width=0.18\linewidth]{fig/uncertainty-progress/#1_#2.png}
    }
    \newcommand{\rowy}[4]{
    \rotatebox{90}{\hspace{1.2em}$\hat{x}_{t \rightarrow 1}$} &
        \imagepng{#1}{x1_#2} & \imagepng{#1}{x1_#3} & \imagepng{#1}{x1_#4} & \imagepng{#1}{img} \\
    \rotatebox{90}{Uncertainty} &
        \imagepng{#1}{#2} & \imagepng{#1}{#3}    & \imagepng{#1}{#4} &
        \multicolumn{1}{@{}l@{}}{
            \includegraphics[height=0.18\linewidth]{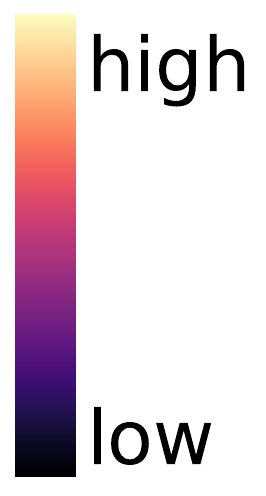}%
      }
      \vspace{0.1cm}
    }
    
    \begin{subfigure}{\linewidth}
    \centering
    \begin{tabular}{ccccc}
        & $t=0.2$ & $t=0.5$ & $t=0.8$ & Image \\
        \rowy{ballon}{t0.20}{t0.50}{t0.80} \\
        \rowy{parrot}{t0.20}{t0.50}{t0.80} \\
    \end{tabular}
    \end{subfigure}
    \vspace{-0.4cm}
    \caption{\textbf{Visualization of one-step predictions at varying timesteps and their corresponding uncertainty maps.} These results show that the model develops a strong intuition about which regions are easy or difficult to generate from very early $t$.
    }
    \vspace{-0.4cm}
    \label{fig:uncertainty-visualization}
\end{figure}

We instead propose a simple alternative that directly constrains the \emph{maximum} information per sample, better aligning training with inference. 
Concretely, we first sample $t_{\max}$, and then conditionally sample $t_i\sim p(t_i \mid t_{\max}; \sigma^2)$, where $p$ is the lower half of a Gaussian distribution centered and truncated at $t_{\max}$ with standard deviation $\sigma$ (\cref{fig:truncate-gaussian-viz}).
Even this simple baseline already improves generation quality over SRM’s uniform-$\bar{t}$ schedule (\cref{fig:srm-comparison}, \textit{Truncated Gaussian}), while also being significantly faster (\cref{fig:our_scheduler_vs_SRM}).

While the truncated Gaussian prevents overly informative training states, sampling $t_{\max} \sim \mathcal{U}(0,1)$ biases the overall timestep distribution toward lower $t$ (i.e., higher noise levels) (\cref{fig:srm-comparison}, middle). To correct this imbalance, we adopt the Logit-Normal timestep sampler from~\cite{esser2024scaling-sdv3}, $\text{LogitNorm}(t; m, s) = \pi(m + s \cdot \epsilon)$ with $\epsilon \sim \mathcal{N}(0, 1)$, $m$ and $s$ being the location and scale parameters, and $\pi$ being the standard logistic function. We term our sampler \emph{Logit-Normal Truncated Gaussian} (LTG), and formalize it as
\begin{align}
    t_{\max} &\sim \text{LogitNorm(m, s)} \\
    t_i &\sim \text{truncate}(\mathcal{N}(t_{\max}, \sigma^2)), \qquad \forall t_i \in \mathbf{t} \nonumber
\end{align}
where the truncate function selects the lower half Gaussian and enforces valid timestep bounds $t \in [0, t_\text{max}]$. Details are provided in \cref{alg:ltg-sampler}. As shown in \cref{fig:srm-comparison}, the LTG sampler yields a more balanced timestep allocation across noise levels and further improves performance.

\paragraph{Estimating Patch-Difficulty} 
To fully leverage \ours{} during sampling, we estimate the relative denoising difficulty of each patch. Inspired by prior works~\cite{seitzerpitfalls, wewer2025spatial_srm, nichol2021improvedDDPM}, we model this difficulty as the predicted standard deviation and minimize the negative log-likelihood (NLL) of the ground-truth conditional velocity $\mathbf{v}_\text{GT} = x_1 - x_0$:
\begin{align}
\mathcal{L}_{\text{total}} &= \mathbb{E} \Bigl[
     \big\| \mathbf{v}_\text{GT} - \mathbf{v}_\theta(\mathbf{x}_t, \mathbf{t}) \big\|^2
     \\[-2pt]
    &- \lambda
    \log \mathcal{N} \bigl(
        \mathbf{v}_{\text{GT}} \mid
            \mathbf{sg}(\mathbf{v}_\theta (\mathbf{x}_t, \mathbf{t})),
            \sigma_\theta(\mathbf{x}_t, \mathbf{t})^2 \mathbf{I}  
    \bigr)
    \Bigr] \nonumber
\end{align}
where $\mathbf{sg}$ is the stop gradient operation and $\lambda$ the NLL loss weight. We follow SRM~\cite{wewer2025spatial_srm} and set $\lambda$ to $0.01$.

\begin{figure}
    \centering
    \includegraphics[width=.9\linewidth]{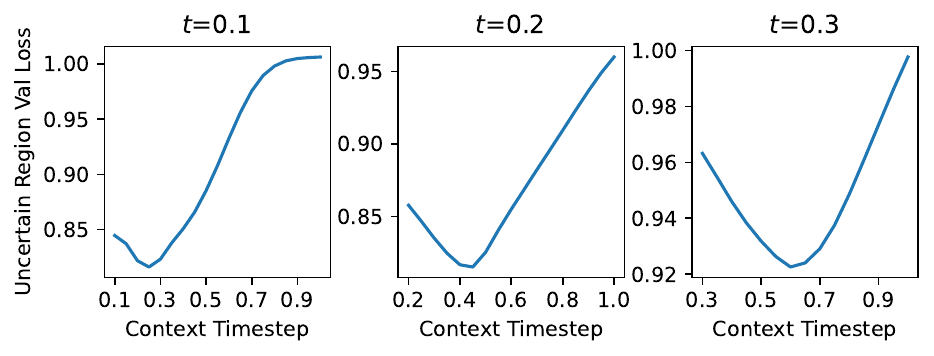}
    \vspace{-0.2cm}
    \caption{\textbf{Validation loss on uncertain regions} with and without additional context from advanced confident patches. Providing "future" context consistently reduces the loss in uncertain areas.
    }
    \label{fig:val-loss-vs-context}
    \vspace{-0.2cm}
\end{figure}

\begin{figure}
    \centering
    \includegraphics[width=.9\linewidth]{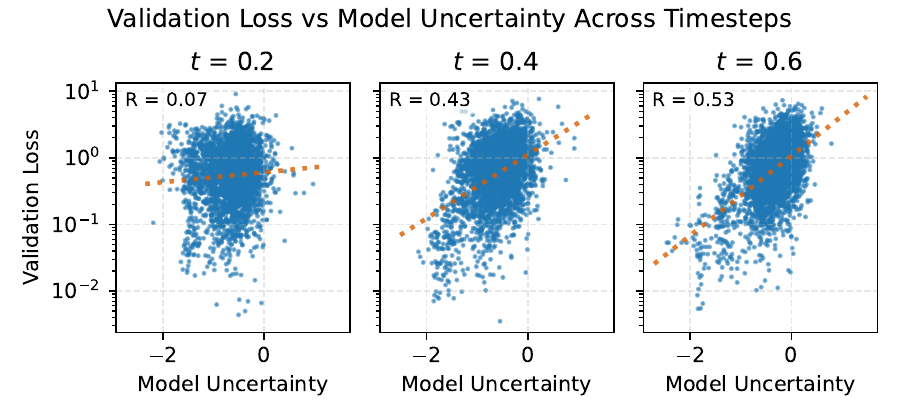}
    \vspace{-0.4cm}
    \caption{\textbf{Correlation of predicted uncertainty and validation loss}: higher uncertainty reliably coincides with higher error.}
    \label{fig:val-loss-vs-uncertainty}
    \vspace{-0.2cm}
\end{figure}

\begin{figure}[htbp]
    \centering
    \setlength{\tabcolsep}{1pt}
    \scriptsize
    \newcommand{\imgw}{0.2\linewidth}
    \begin{tabular}{ccccc}
        Image & {unc mask} & {unc w/o context} & {unc w/ context} \\
        
        \includegraphics[width=\imgw]{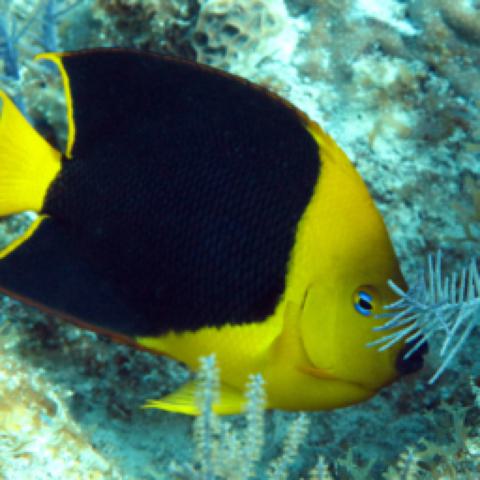} &
        \includegraphics[width=\imgw]{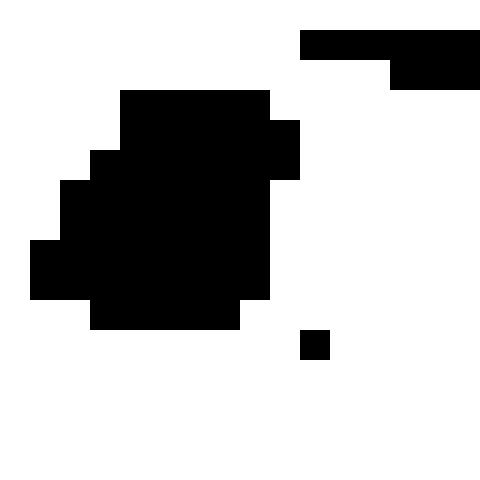} &
        \includegraphics[width=\imgw]{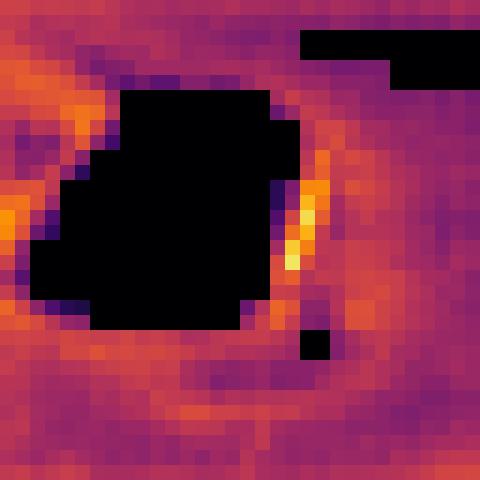} &
        \includegraphics[width=\imgw]{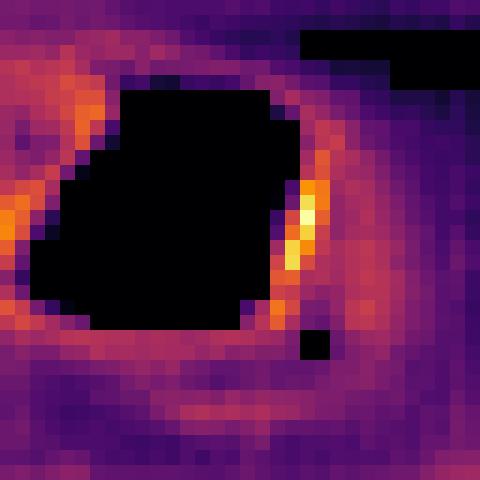} &
        \includegraphics[height=\imgw]{fig/uncertainty-progress/0_colorbar_vertical.pdf}\\
        
        \includegraphics[width=\imgw]{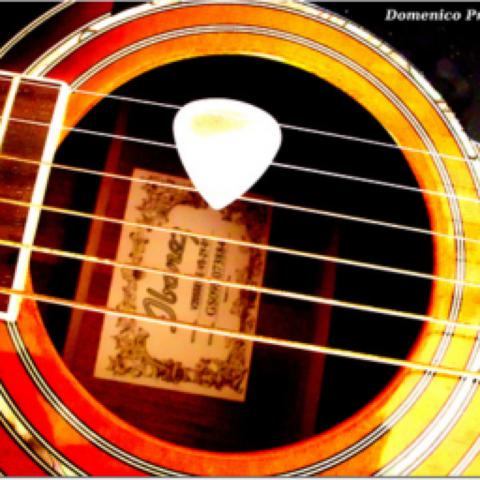} &
        \includegraphics[width=\imgw]{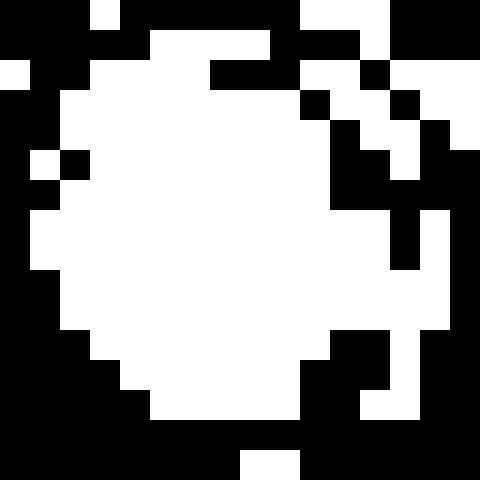} &
        \includegraphics[width=\imgw]{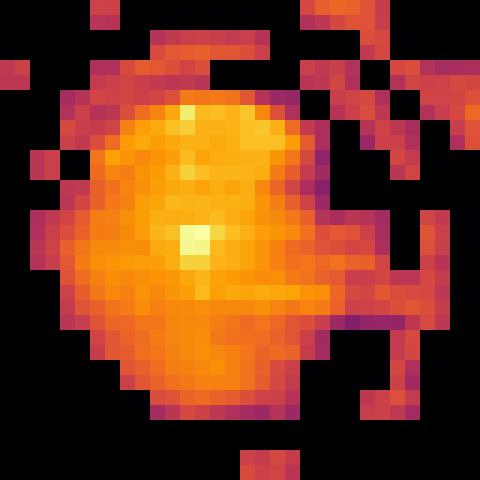} &
        \includegraphics[width=\imgw]{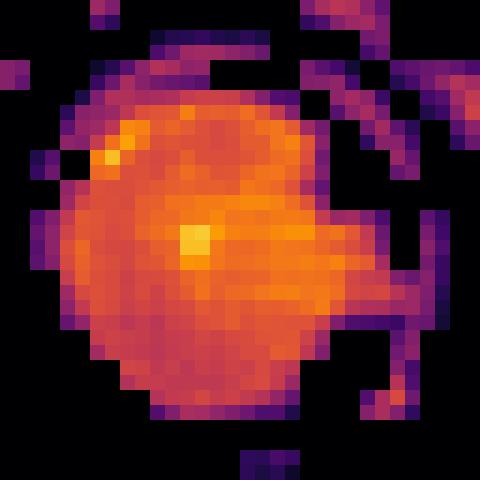} &
        \includegraphics[height=\imgw]{fig/uncertainty-progress/0_colorbar_vertical.pdf}\\
    \end{tabular}
    \vspace{-0.2cm}
    \caption{\textbf{Effect of context on predicted uncertainty.} When confident regions are advanced and provide context, the predicted uncertainty in the remaining high-uncertainty regions decreases, evaluated at the same timestep $t$.}
    \label{fig:uncertainty-vs-context-viz}
    \vspace{-0.2cm}
\end{figure}

\subsection{Adaptive Sampling}
\label{sec:pf:sampling}
Combining uncertainty prediction with varying noise scales across patches offers our model particularly flexible sampling strategies.
Beyond the standard parallel sampling, we explore adaptive sampling strategies that explicitly leverage predicted patch difficulty. To understand the role of the predicted uncertainty, we first analyze its behavior. We visualize the model’s predicted uncertainty across different timesteps $t$ in \cref{fig:uncertainty-visualization} and observe that it assigns higher uncertainty to semantically or structurally challenging regions.
We further find that it strongly correlates with validation error (\cref{fig:val-loss-vs-uncertainty}). Regions with low predicted uncertainty correspond to areas where the model exhibits higher confidence and performs more accurate few-step predictions.

We then examine in \cref{fig:val-loss-vs-context} whether uncertain regions can benefit from additional context provided by high-confidence regions predicted by the model. We compare two settings: \textit{a)} both uncertain and confident regions are evaluated at the same timestep $t$; and \textit{b)} the confident regions advance by a small step $\delta t$ using one-step prediction, effectively providing ``future'' context. Our results show that this contextual conditioning results in lower validation loss in uncertain regions. 
Importantly, as briefly discussed in \cref{sec:related}, this effect is different and orthogonal to guidance techniques~\cite{ho2021classifier_cfg, dhariwal2021diffusion}. Our method modifies the conditioning state by exposing uncertain patches to a locally more resolved representation generated by the model itself. This form of context conditioning remains fully self-consistent with the training and avoids out-of-distribution samples.

Given our patch forcing framework and the model’s ability to predict patch-wise uncertainty, we can derive adaptive sampling strategies that dynamically allocate compute.
To this end, we introduce two adaptive, uncertainty-guided samplers that dynamically identify confident regions from predicted uncertainty maps and propagate them forward in time to provide context for the remaining regions. Both samplers identify confident regions using a threshold on the predicted uncertainty. We visualize these samplers in \cref{fig:schedule-vs-fid}.

\paragraph{Dual-Loop} We term our first sampler \textit{dual-loop} as it preferentially advances confident regions to provide context for uncertain ones, while also refining the latter with more steps. At each iteration, we (i) take larger timestep updates on low-uncertainty patches (outer loop) and (ii) condition on their updated states to denoise high-uncertainty patches with smaller steps (inner loop). Once both subsets align at the same timestep, we re-estimate the uncertainty and repeat, allowing the patch difficulties to adapt over time.

\paragraph{Look-Ahead} Our second sampler is called \textit{look-ahead} as propagates contextual representations of confident regions to a future timestep $ t$ proportional to the current $t$, up to the clean data $t=1$. These advanced representations then serve as contextual guidance for denoising the more uncertain regions. We detail the proposed sampling procedure in \cref{alg:fm-context-transport} and show examples in \cref{fig:schedule-vs-fid}.

\begin{figure}[t]
    \centering \small
    \setlength\tabcolsep{1pt}

    \newcommand{\baseimg}[1]{%
        \includegraphics[width=0.15\linewidth]{fig/uncertainty-noise/image_#1.jpg}}
    \newcommand{\singlestep}[1]{%
        \includegraphics[width=0.15\linewidth]{fig/uncertainty-noise/single_step_std_#1.jpg}}
    \newcommand{\uncertaintyimg}[2]{%
        \includegraphics[width=0.15\linewidth]{fig/uncertainty-noise/uncertainty_#1_#2.jpg}}

    \newcommand{\onerow}[1]{%
        \baseimg{#1} & \singlestep{#1} &
        \uncertaintyimg{1}{#1} & \uncertaintyimg{2}{#1} &
        \uncertaintyimg{4}{#1} & \uncertaintyimg{6}{#1} \\}

    \begin{tabular}{cccccc}
        Image & Var($x_{t\rightarrow 1}$) & unc $\mathit{1}$ & unc $\mathit{2}$ & unc $\mathit{3}$ & unc $\mathit{4}$ \\
        \onerow{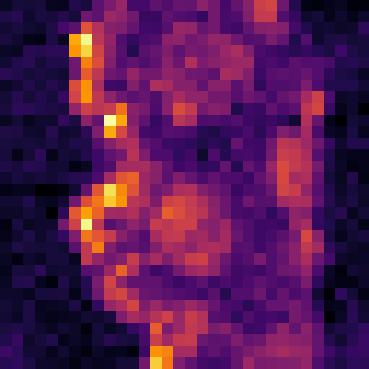}
        \onerow{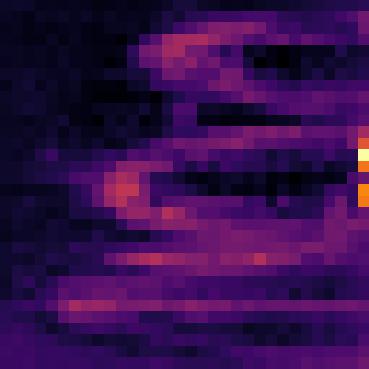}
    \end{tabular}
    \vspace{-0.2cm}
    \caption{
    \textbf{Patch difficulty prediction} aligns with the model's $x_1$ prediction variance.
    }
    \label{fig:uncertainty-noise-alignment}
    \vspace{-0.3cm}
\end{figure}

\section{Experiments}
\label{sec:experiments}
\begin{table}[t]
\centering
\footnotesize
\begin{tabular}{lrrl}
  \toprule
  \textbf{Model} & \textbf{\#Param} & \textbf{Iter.} & \textbf{FID} $\downarrow$ \\
  \midrule
  SiT-B/2 \cite{ma2024sit} & \multirow{4}{*}{130M} & \multirow{4}{*}{400K} & 33.0 \\
  PFT-B/2 &  &  & 27.9 \\
  \hspace{1em} + dual-loop &  &  & 26.0 \\
  \hspace{1em} + look-ahead &  &  & \textbf{24.2} \\
  \lightmidrule
  SiT-L/2 \cite{ma2024sit} & \multirow{4}{*}{458M} & \multirow{4}{*}{400K} & 18.8 \\
  PFT-L/2 &  &  & 14.7 \\
  \hspace{1em} + dual-loop &  &  & 13.9 \\
  \hspace{1em} + look-ahead &  &  & \textbf{13.0} \\
  \lightmidrule
  SiT-XL/2 \cite{ma2024sit} & \multirow{5}{*}{675M} & \multirow{5}{*}{400K} & 17.2 \\
  \hspace{1em} + REPA & & & 7.9 \\
  PFT-XL/2 &  &  & 12.9 \\
  \hspace{1em} + dual-loop &  &  & 11.5 \\
  \hspace{1em} + look-ahead &  &  & 9.8 \\
  \hspace{1em} + look-ahead + REPA &  &  & \textbf{6.7} \\
  \bottomrule
\end{tabular}
\vspace{-0.2cm}
\caption{\textbf{Performance comparison on ImageNet $256^2$.} With fixed architecture and NFE, our Patch Forcing Transformer (PFT) already outperforms the baseline under standard Euler sampling. Our proposed dual-loop and look-ahead samplers further improve performance by allocating function evaluations more effectively.
}
\label{tab:dit}
\vspace{-0.1cm}
\end{table}

\begin{table}[t]
\centering
\setlength{\tabcolsep}{6pt}
\adjustbox{max width=\linewidth}{
\begin{tabular}{lcccccc}
\toprule
Model & Epochs & FID$\downarrow$ & sFID$\downarrow$ & IS$\uparrow$ & Pre.$\uparrow$ & Rec.$\uparrow$ \\
\midrule
MaskDiT~\cite{zheng2023maskdit}
    & 1600 & 2.28 & 5.67 & 276.6 & 0.80 & 0.61 \\
SD-DiT~\cite{zhu2024sd-dit}
    &  480 & 3.23 &   -  &   -   &   -  &   -  \\
DiT-XL/2~\cite{peebles2023scalable-dit}
    & 1400 & 2.27 & 4.60 & 278.2 & \textbf{0.83} & 0.57 \\
SiT-XL/2~\cite{ma2024sit}
    & 1400 & 2.06 & 4.50 & 270.3 & 0.82 & 0.59 \\
SiT-XL/2 + REPA~\cite{yurepresentation_repa}
    &  \textbf{200} & \textbf{1.96} & 4.49 & 264.0 & 0.82 & 0.60 \\
PFT-XL/2 + REPA + look-ahead & \textbf{200} & 2.00 & \textbf{4.32} & \textbf{284.1} & 0.81 & \textbf{0.61} \\
\bottomrule
\end{tabular}
}
\vspace{-0.2cm}
\caption{\textbf{State-of-the-Art comparison on ImageNet} $256^2$.}
\label{tab:dit-cfg}
\vspace{-0.3cm}
\end{table}

We first evaluate our method on class-conditional ImageNet~\cite{russakovsky2015imagenet} at $256\times256$ resolution, and later show that it also extends to text-to-image synthesis. For ImageNet, we keep the backbone architecture, and thus the number of parameters, fixed (SiT/DiT variants B, L, and XL). Timestep conditioning in DiT's follows AdaLN~\cite{peebles2023scalable-dit}. Extending it to different timesteps per patch is straightforward: at each denoising step, we provide a per-token timestep embedding by sampling one timestep per token, which requires only minimal architectural changes (see \Cref{fig:supp:code-sit}). The only additional parameters arise from the uncertainty prediction head, which imposes minimal overhead (less than 0.01\%).

\begin{figure}
    \centering
    \includegraphics[width=0.99\linewidth]{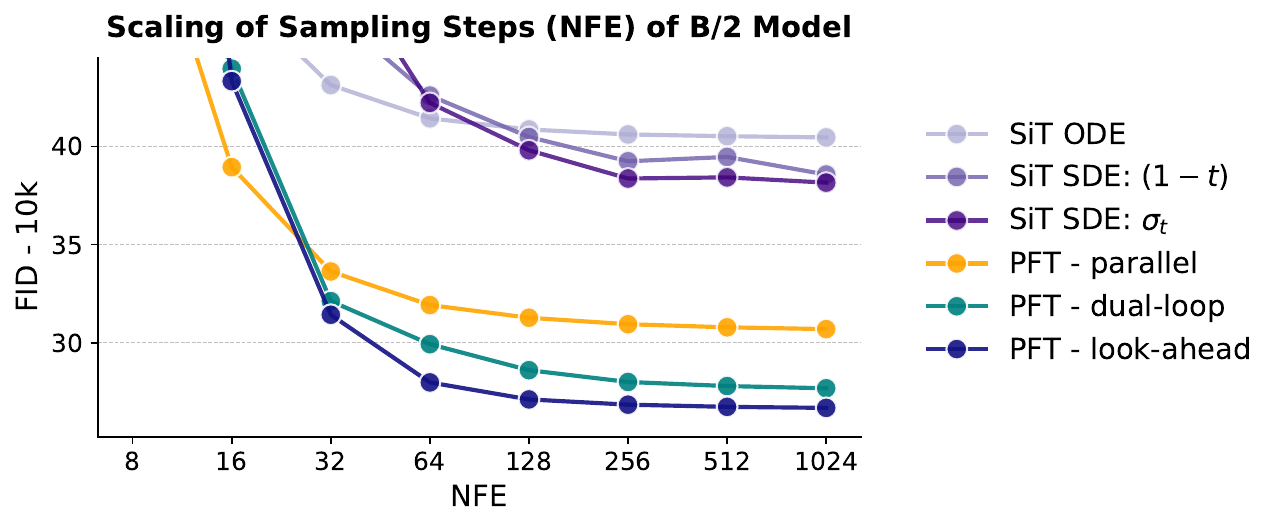}
    \vspace{-0.2cm}
    \caption{\textbf{Scaling sampling steps.} Across increasing numbers of function evaluations (NFE), our PFT-B/2 model consistently outperforms the SiT-B/2 ODE and SDE baselines. Our uncertainty-aware samplers further improve over parallel PFT sampling, with dual-loop and look-ahead achieving the best FID across NFEs.
    }
    \label{fig:nfe-fid}
    \vspace{-0.2cm}
\end{figure}

\begin{figure}
    \centering
    \includegraphics[width=0.9\linewidth]{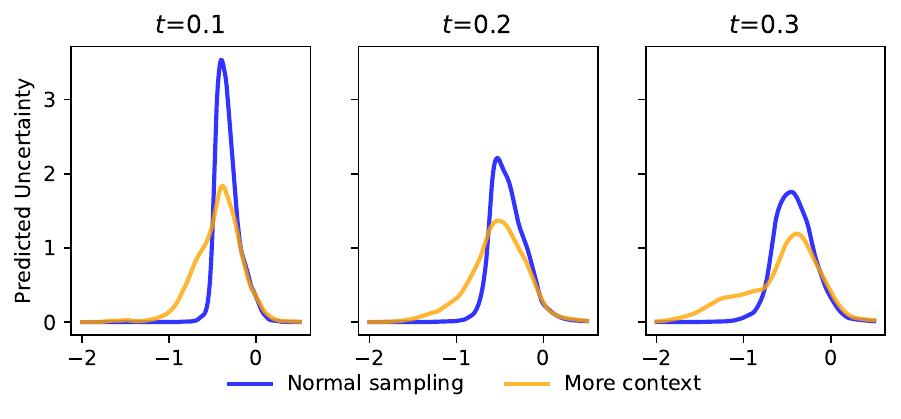}
    \vspace{-0.2cm}
    \caption{\textbf{More context reduces uncertainty.} Predicted uncertainty aligns with our qualitative findings in \cref{fig:val-loss-vs-context}: as additional context is explicitly introduced through the confident regions, the uncertainty of the remaining regions consistently decreases.
    } %
    \label{fig:uncertainty-vs-context}
    \vspace{-0.3cm}
\end{figure}

\subsection{Class-conditional Image Synthesis}
In~\cref{tab:dit} we report FID-50k on ImageNet~\cite{russakovsky2015imagenet}. With fixed architecture and compute, our Patch Forcing Transformer (PFT) with parallel Euler sampling already improves over SiT, and these gains transfer across model scales. Adding our uncertainty-aware samplers further boosts performance; in particular, the \emph{look-ahead} sampler delivers the strongest improvements, indicating that exposing uncertain patches to lower-noise context helps generation.
In addition, \cref{tab:dit} shows that pairing PFT-XL/2 with REPA yields comparable relative gains and improves over the SiT baseline in both settings, indicating that Patch Forcing is orthogonal to representation alignment. Likewise, \cref{tab:dit-cfg} confirms orthogonality to classifier-free guidance (CFG): combining CFG with our look-ahead sampler achieves superior or competitive performance to prior state-of-the-art.
The scaling curves in \cref{fig:nfe-fid} further show that our samplers also improve with more sampling steps. Notably, our adaptive samplers improve over both ODE and SDE sampling by a large margin. We show qualitative results in \cref{fig:supp:qualitative-imagenet}.

\subsection{Improved Timestep Sampler}
We next study how to sample per-patch timesteps during training. As shown in \cref{fig:srm-comparison}, the Uniform $\bar{t}$ scheduler from Spatial Reasoning Models (SRM)~\cite{wewer2025spatial_srm} balances the marginal distributions of both the individual $t_i$ and their mean $\bar{t}$, yet the per-sample maximum timestep $t_{\max}$ remains relatively high. This implies that each training sample still contains partially denoised context, creating a train–test mismatch for image synthesis, where inference begins from pure noise. The gap is reflected quantitatively: across three models trained with different $\beta$-sharpness values, where larger $\beta$ approaches parallel timestep sampling, $\beta{=}1$ (widest spread) performs worst. Only at higher sharpness results surpass the SiT baseline. We also evaluate a truncated Gaussian schedule, which limits $t_{\max}$ and improves over SRM but collapses the $t$ mass toward small values, severely reducing timestep variety. Our proposed Logit-Normal Truncated Gaussian (LTG) timestep sampler addresses both issues, spreading the individual $t_i$ while controlling $t_{\max}$; correspondingly, it yields the best FID among the evaluated schedules.

\subsection{Difficulty-Aware Samplers}
As discussed in \cref{sec:pf:sampling}, the uncertainty head enables the design of flexible and adaptive sampling strategies. In \cref{fig:schedule-vs-fid} left, we visualize two such approaches: the dual-loop sampler first advances confident pixels with larger steps while using smaller steps for uncertain ones; and the look-ahead sampler, which explicitly propagates confident pixels to a future state to provide contextual guidance.

\Cref{fig:schedule-vs-fid} (right), shows a quantitative comparisons across different schedulers under fixed number of function evaluations (NFE). Our Patch Forcing Transformer (PFT) consistently outperforms the baseline across all scenarios, confirming the benefit of patch-wise adaptivity. Moreover, we show that patch ordering plays a critical role: in the PFT-random variant, where context is propagated from randomly selected pixels, performance degrades compared to the PFT-parallel baseline, highlighting the importance of uncertainty-informed patch prioritization. We ablate and analyze these samplers in more detail in the Appendix.

\subsection{Validating the Three Key Findings}
\paragraph{Context helps generation.}
As shown in \cref{fig:val-loss-vs-context}, providing additional context to uncertain regions reduces their validation loss. When confident patches are advanced to lower-noise states and used as context, while evaluating error on the remaining uncertain regions at the same $t$, the reconstruction error decreases. Note that the optimal amount of context is proportional to timestep $t$. This observation motivates the design of our look-ahead sampler, which advances confident regions proportionally into the future to provide appropriate context. This effect is robust across timesteps.

\paragraph{Uncertainty indicates patch difficulty.} \cref{fig:val-loss-vs-uncertainty} shows that the per-patch predicted uncertainty is positively correlated with the corresponding validation loss. The fitted regression (orange) confirms this relationship, with later timesteps exhibiting stronger correlation (e.g., $R=0.52$ at $t=0.6$) than early, near-noise timesteps (e.g., $R=0.11$ at $t=0.2$). This trend indicates that the uncertainty signal becomes more diagnostic as denoising progresses.
We further assess this correlation qualitatively in \cref{fig:uncertainty-noise-alignment}. To test whether this signal reflects intrinsic patch denoising difficulty, we fix an image, corrupt it to $x_t$, predict the per-patch uncertainty, and perform one-step predictions toward $x_{t\rightarrow 1}$ across multiple noise realizations, measuring the per-patch empirical variance via Monte Carlo sampling. Predicted uncertainty strongly aligns with the ensemble variance, indicating that the estimates reliably capture relative regional difficulty.

\paragraph{More context reduces uncertainty.} \Cref{fig:uncertainty-vs-context-viz} visualizes that injecting explicit context from confident patches lowers the predicted uncertainty of the remaining regions when evaluated at the same timestep. Quantitatively, \cref{fig:uncertainty-vs-context} shows the uncertainty histogram shifting toward smaller values after look-ahead, mean uncertainty decreases, and mass moves away from the high-uncertainty tail, confirming that localized, high-fidelity context actively resolves ambiguity rather than merely correlating with it.

\subsection{Scaling Patch Forcing to T2I}
We evaluate the generalization of our approach beyond class-conditional ImageNet by scaling it to text-conditional generation. Specifically, we train a 1.2B Patch Forcing transformer (PFT) with Qwen3-1.7B~\cite{qwen3vlembedding} as text encoder on a subset of 120M image–text pairs from COYO~\cite{kakaobrain2022coyo-700m} recaptioned with InternVL3-2B~\cite{zhu2025internvl3} and evaluate it on T2I-CompBench++\cite{huang2023t2i} and GenEval~\cite{ghosh2023genevalobjectfocusedframeworkevaluating}.
\cref{tab:supp:t2i-compbench-baselines} and \cref{tab:supp:t2i-geneval-baselines} show that our approach scales effectively to text-to-image synthesis and achieves competitive results across benchmarks. At fixed NFE, our difficulty-aware sampler denoises more effectively than standard Euler sampling, showing that the benefits of Patch Forcing are not limited to class-conditional generation.
We further observe that, compared to a similarly trained model using standard Flow Matching, Patch Forcing produces clearer text rendering (see \Cref{fig:t2i-spelling}). We analyze this effect further in Appendix \cref{sec:supp:t2i} and provide additional qualitative samples in \Cref{fig:supp:qualitative-text-rendering}.

\begin{table}[t]
\centering
\resizebox{\linewidth}{!}{
\begin{tabular}{lcccccc}
\toprule
& \makecell{\textbf{Color} \\ B-VQA}
& \makecell{\textbf{Shape} \\ B-VQA}
& \makecell{\textbf{Texture} \\ B-VQA}
& \makecell{\textbf{2D-Spatial} \\ UniDet}
& \makecell{\textbf{3D-Spatial} \\ UniDet}
& \makecell{\textbf{Non-Spatial} \\ CLIP} \\
\midrule
SDv1.4~\cite{rombach2022high-sd-ldm} & 0.3765 & 0.3576 & 0.4156 & 0.1246 & 0.3030 & 0.3079 \\
SDv2~\cite{rombach2022high-sd-ldm}    & 0.5065 & 0.4221 & 0.4922 & 0.1342 & 0.3230 & 0.3096 \\
Composable v2~\cite{liu2023compositionalvisualgenerationcomposable} & 0.4063 & 0.3299 & 0.3645 & 0.0800 & 0.2847 & 0.2980 \\
Structured v2~\cite{feng2023trainingfreestructureddiffusionguidance} & 0.4990 & 0.4218 & 0.4900 & 0.1386 & 0.3224 & 0.3111 \\
Attn-Exct v2~\cite{chefer2023attend} & 0.6400 & 0.4517 & 0.5963 & 0.1455 & 0.3222 & 0.3109 \\
GORS~\cite{huang2023t2i} & 0.6603 & 0.4785 & 0.6287 & 0.1815 & 0.3572 & 0.3193 \\
Dalle-2~\cite{ramesh2022hierarchical-dalle2} & 0.5750 & 0.5464 & 0.6374 & 0.1283 & - & 0.3043 \\
SDXL~\cite{podellsdxl} & 0.6369 & 0.5408 & 0.5637 & 0.2032 & - & 0.3110 \\
PixArt-$\alpha$~\cite{chen2023pixartalpha} & 0.6886 & 0.5582 & 0.7044 & 0.2082 & - & 0.3179 \\
\midrule
PFT-1.2B & 0.7132 & 0.4951 & 0.6316 & 0.1903 & 0.2770 & 0.3034 \\
\hspace{1em} + dual-loop & 0.7058 & 0.4947 & 0.6437 & 0.1956 & 0.2771 & 0.3034\\
\hspace{1em} + look-ahead & 0.7323 & 0.5036 & 0.6444 & 0.1956 & 0.2815 & 0.3035  \\
\bottomrule
\end{tabular}
}
\vspace{-0.2cm}
\caption{\textbf{T2I-CompBench++} evaluation~\cite{huang2023t2i}.}
\label{tab:supp:t2i-compbench-baselines}
\vspace{-0.2cm}
\end{table}

\begin{table}[t]
\centering
\resizebox{\linewidth}{!}{
\begin{tabular}{lccccccc}
\toprule
& \textbf{Single Obj.} & \textbf{Two Obj.} & \textbf{Counting} & \textbf{Colors} & \textbf{Position} & \textbf{Color Attri.} & \textbf{Overall}  \\
\midrule
LlamaGen~\cite{sun2024llamagen} & 0.71 & 0.34 & 0.21 & 0.58 & 0.07 & 0.04 & 0.32 \\
LDM~\cite{rombach2022high-sd-ldm} & 0.92 & 0.29 & 0.23 & 0.70& 0.02 & 0.05 & 0.37 \\
SDv1.5~\cite{rombach2022high-sd-ldm}  & 0.97 & 0.38 & 0.35 & 0.76 & 0.04 & 0.06 & 0.43 \\
PixArt-$\alpha$~\cite{chen2023pixartalpha} & 0.98 & 0.50 & 0.44 & 0. 80 & 0.08 & 0.07 & 0.48 \\
SDv2.1~\cite{rombach2022high-sd-ldm}  & 0.98 & 0.51 & 0.44 & 0.85 & 0.07 & 0.17 & 0.50 \\
DALL-E2~\cite{ramesh2022hierarchical-dalle2} & 0.94 & 0.66 & 0.49 & 0.77 & 0.10 & 0.19 & 0.52 \\
Emu3-Gen~\cite{wang2024emu3nexttokenpredictionneed} & 0.98 & 0.71 & 0.34 & 0.81 & 0.17 & 0.21 & 0.54 \\
SDXL~\cite{podellsdxl} & 0.98 & 0.74 & 0.39 & 0.85 & 0.15 & 0.23 & 0.55 \\
DALL-E3~\cite{dalle3} & 0.96 & 0.87 & 0.47 & 0.83 & 0.43 & 0.45 & 0.67 \\
SD3-Medium~\cite{esser2024scaling-sdv3} & 0.99 & 0.94 & 0.72 & 0.89 & 0.33 & 0.60 & 0.74 \\
\midrule
PFT-1.2B & 0.94 & 0.66 & 0.44 & 0.76 & 0.32 & 0.45 & 0.59 \\
\hspace{1em} + dual-loop & 0.94 & 0.69 & 0.43 & 0.70 & 0.36 & 0.40 & 0.59 \\
\hspace{1em} + look-ahead & 0.97 & 0.73 & 0.41 & 0.86 & 0.35 & 0.46 & 0.63 \\
\bottomrule
\end{tabular}
}
\vspace{-0.2cm}
\caption{\textbf{T2I GenEval} evaluation~\cite{ghosh2023genevalobjectfocusedframeworkevaluating}.}
\label{tab:supp:t2i-geneval-baselines}
\vspace{-0.2cm}
\end{table}

\begin{figure}[htbp]
  \centering
  \setlength{\tabcolsep}{2pt}
  \begin{tabular}{ccc}
    \includegraphics[width=0.3\linewidth]{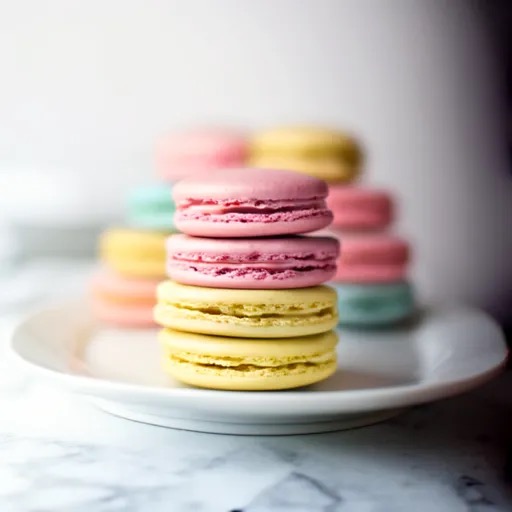} &
    \includegraphics[width=0.3\linewidth]{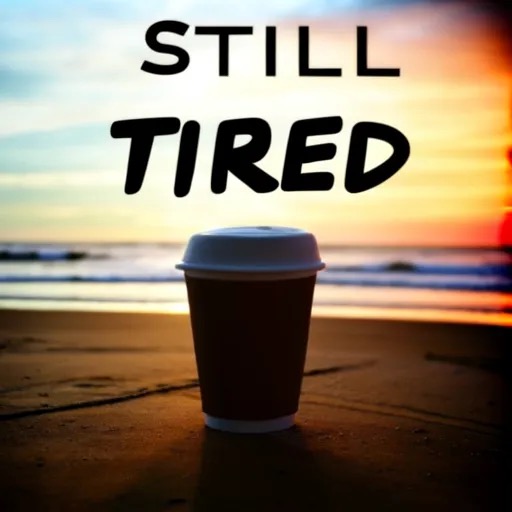} &
    \includegraphics[width=0.3\linewidth]{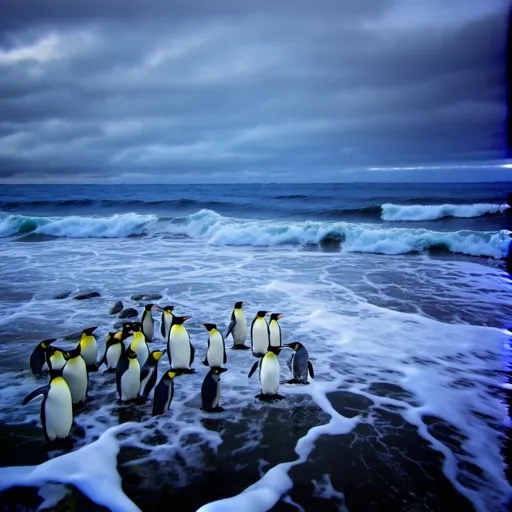}
    \\
    \includegraphics[width=0.3\linewidth]{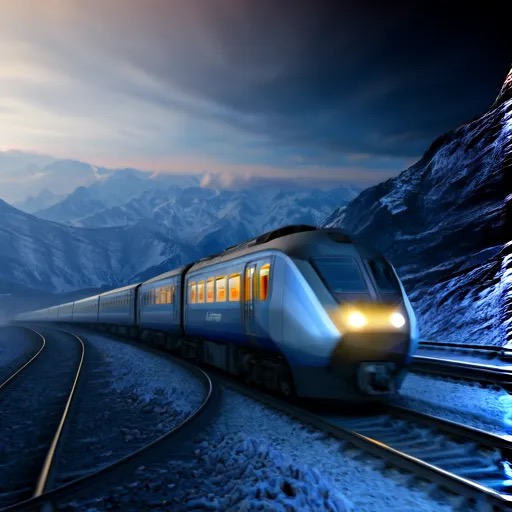} &
    \includegraphics[width=0.3\linewidth]{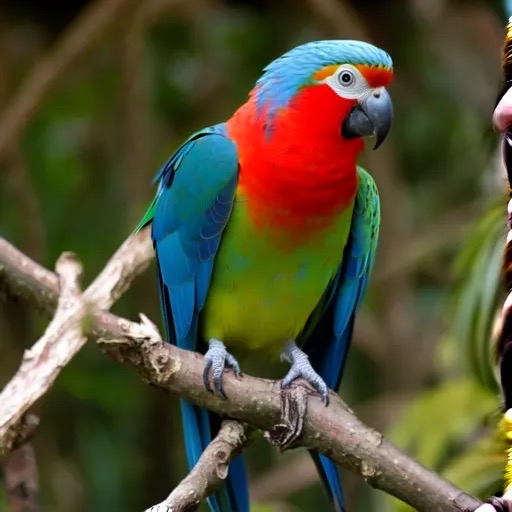} &
    \includegraphics[width=0.3\linewidth]{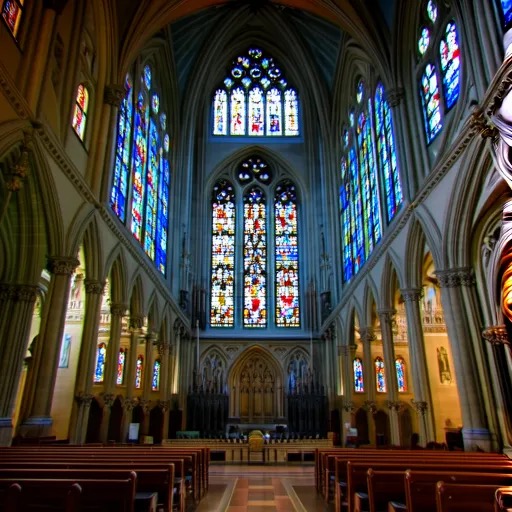}
  \end{tabular}
  \vspace{-0.3cm}
  \caption{\textbf{Qualitative text-to-image} results at 512 px resolution.}
  \label{fig:t2i-grid}
  \vspace{-0.2cm}
\end{figure}

\begin{figure}[htbp]
  \centering
  \small
  \setlength{\tabcolsep}{2pt}
  \begin{tabular}{c c c c}
    \rotatebox{90}{\hspace{1.4em} Baseline} &
    \includegraphics[width=0.28\linewidth]{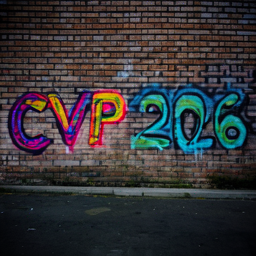} &
    \includegraphics[width=0.28\linewidth]{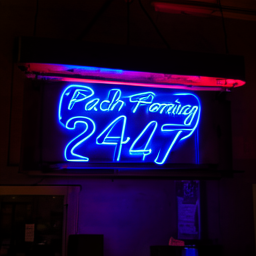} &
    \includegraphics[width=0.28\linewidth]{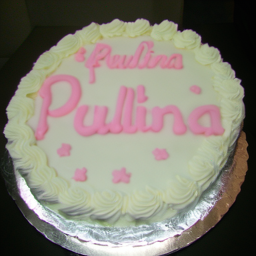}
    \\
    \rotatebox{90}{\hspace{0.7em} Patch Forcing} &
    \includegraphics[width=0.28\linewidth]{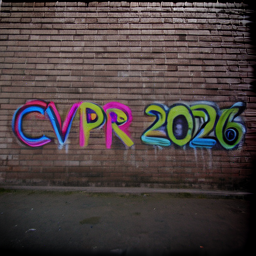} &
    \includegraphics[width=0.28\linewidth]{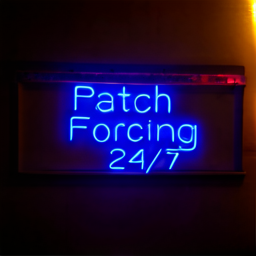} &
    \includegraphics[width=0.28\linewidth]{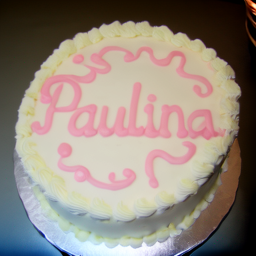}
    \\
    & \parbox{0.25\linewidth}{\centering\tiny Graffiti on a brick wall spelling “CVPR 2026” in colorful font.}
    & \parbox{0.25\linewidth}{\centering\tiny A neon sign over a bar that reads "Patch-Forcing 24/7" with blue font.}
    & \parbox{0.25\linewidth}{\centering\tiny A birthday cake with icing that spells "Paulina" in pink font.}
  \end{tabular}
  \caption{\textbf{Text rendering comparison.} Our PFT shows superior text rendering compared to an equivalent model trained with vanilla Flow Matching, under identical training and inference settings (plain Euler sampler, fixed NFE, same seed). Additional uncurated samples in \Cref{fig:supp:qualitative-text-rendering}.}
  \vspace{-0.5cm}
  \label{fig:t2i-spelling}
\end{figure}

\section{Conclusion}
\label{sec:conclusion}
We introduce Patch Forcing (PF), a simple and flexible framework for spatially adaptive image synthesis based on per-patch timesteps and predicted patch difficulty. By allowing different regions of an image to follow different noise trajectories, PF enables confident regions to move ahead and provide useful context for more challenging ones.
We show that patch-level timestep schedules already improve image generation when paired with an appropriate training timestep sampler, and that combining them with difficulty-aware samplers yields further gains. Across ImageNet and text-to-image benchmarks, PF consistently outperforms strong baselines and remains compatible with existing guidance methods, such as REPA and CFG. Overall, our results suggest that patch-level timesteps and denoising schedules are a promising foundation for a new class of adaptive samplers that allocate compute where it is most needed.
A natural direction is to extend this idea to few-step and distilled models, where larger time jumps may amplify the benefits of adaptive patch-wise progression.

\onecolumn
\twocolumn
\section*{Acknowledgments}
We would like to thank Nick Stracke and Kolja Bauer for helpful discussions, Jannik Wiese for assistance with design, and Owen Vincent for technical support.
This project has been supported by the bidt project KLIMA-MEMES, the Horizon Europe project ELLIOT (GA No.\ 101214398), the project ``GeniusRobot'' (01IS24083) funded by the Federal Ministry of Research, Technology and Space (BMFTR), the BMWE ZIM-project (No.\ KK5785001LO4) ``conIDitional LoRA'', and the German Federal Ministry for Economic Affairs and Energy within the project ``NXT GEN AI METHODS - Generative Methoden für Perzeption, Prädiktion und Planung''. The authors gratefully acknowledge the Gauss Center for Supercomputing for providing compute through the NIC on JUWELS/JUPITER at JSC and the HPC resources supplied by the NHR@FAU Erlangen.

\section*{Author Contributions}
JS led the project, developed the core methodology, and implemented the initial prototype. JS and MG jointly implemented the framework. JS conducted the ImageNet experiments and optimized the text-to-image models. MG contributed to sampler development and analyzed the models and caching mechanisms. FK conducted the data preparation. YL and FK evaluated the text-to-image models. PM and FK contributed to discussions and related work. BO supervised the project and reviewed the manuscript. All authors contributed to writing.

{
    \small
    \bibliographystyle{ieeenat_fullname}
    \bibliography{main}
}

\appendix
\clearpage
\setcounter{page}{1}
\maketitlesupplementary

\setcounter{figure}{0}
\renewcommand{\thefigure}{S\arabic{figure}}

\setcounter{table}{0}
\renewcommand{\thetable}{S\arabic{table}}

\setcounter{algorithm}{0}
\renewcommand{\thealgorithm}{S\arabic{algorithm}}

\begin{figure}
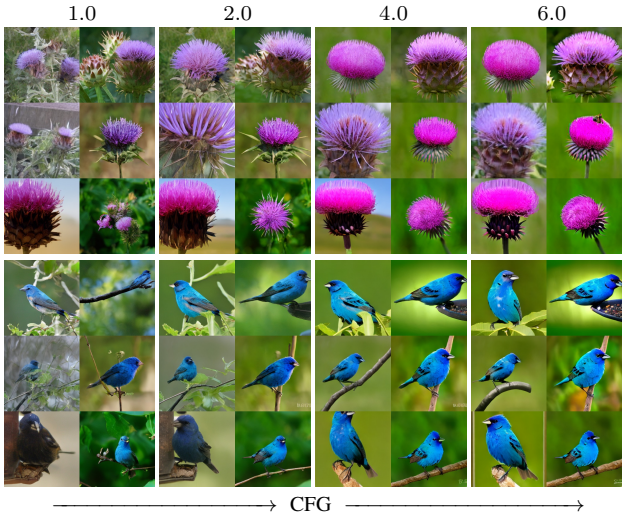

    \footnotesize
    \setlength\tabcolsep{0pt}
    \newcommand{\imagepng}[2]{
        \includegraphics[width=0.24\linewidth]{fig/supp/cfg-abl/repa-pf-xl_cls#1_cfg#2.jpeg}
    }
    \newcommand{\rowy}[1]{
        \imagepng{#1}{1.0} & \imagepng{#1}{2.0} &
        \imagepng{#1}{4.0} & \imagepng{#1}{6.0}
    }
    \centering
    \begin{tabular}{cccc}
        $1.0$ & $2.0$ & $4.0$ & $6.0$ \\
        \rowy{946-cardoon_step25} \\
        \rowy{14-indigo-bunting_step25} \\
        \multicolumn{4}{c}{
            $\xrightarrow{\hspace{2.8cm}}\;
            \text{CFG}\;
            \xrightarrow{\hspace{3cm}}$
        } \\
    \end{tabular}
    \vspace{-0.2cm}
    \caption{\textbf{Uncurated Imagenet $256 \times 256$ samples} with increasing CFG scale from our REPA-PF-XL model. We use $25$ outer steps and $4$ inner steps, resulting in a total of $100$ NFE with our \textit{dual-loop} sampler.}
    \label{fig:supp:cfg-ablation}
\end{figure}

\section{Further Results}

\subsection{Class-conditional Generation}

\begin{figure}
    \centering
    \includegraphics[width=0.8\linewidth]{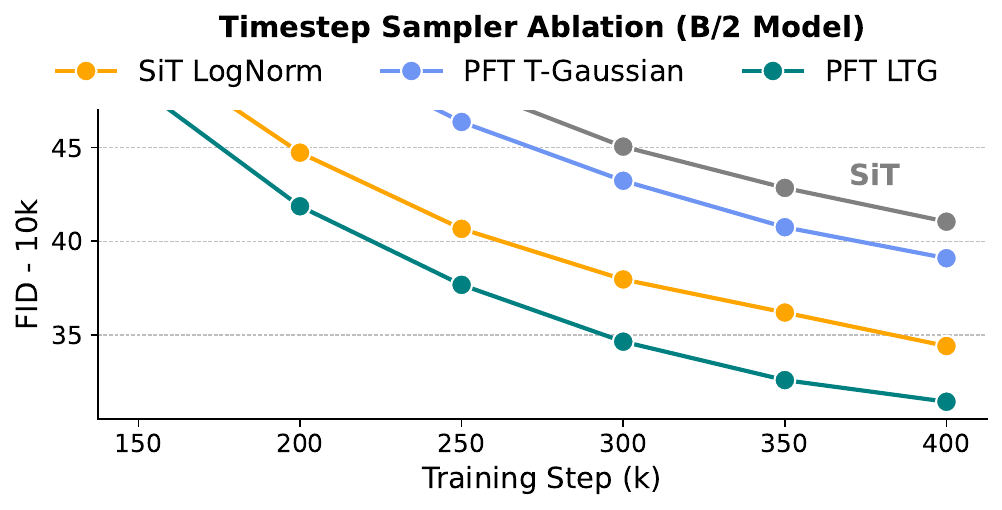}
    \caption{\textbf{Ablation of timestep samplers} under a fixed NFE budget of $100$. Our proposed PFT with Logit-Normal Truncated Gaussian (LTG) shows orthogonal gains and consistently outperforms the SiT Logit-Normal~\cite{esser2024scaling-sdv3} and the Gaussian samplers.
    }
    \label{fig:supp:abl-t-sampler}
\end{figure}

\paragraph{Comparison of Timestep Schedulers}
We compare different timestep schedulers in \cref{fig:supp:abl-t-sampler} to identify the source of the gains from our PFT Logit-Normal Truncated Gaussian (LTG) sampler. To disentangle the individual effects, we compare against the SiT baseline with Logit-Normal timestep sampling, verifying that the improvement is not solely due to the Logit-Normal schedule, and against a Patch Forcing model with a simple truncated Gaussian sampler, verifying that the gain is not merely due to Patch Forcing. PFT with LTG consistently outperforms both baselines, showing that the benefits of Logit-Normal timestep sampling and truncated Gaussian patch-wise timestep allocation are complementary.

\begin{figure}
    \centering
    \definecolor{sit}{HTML}{808080}
    \definecolor{repa}{HTML}{ffa500}
    \definecolor{euler}{HTML}{6aadd5}
    \definecolor{dualloop}{HTML}{3787c0}
    \definecolor{lookahead}{HTML}{08306b}
    \includegraphics[width=0.8\linewidth]{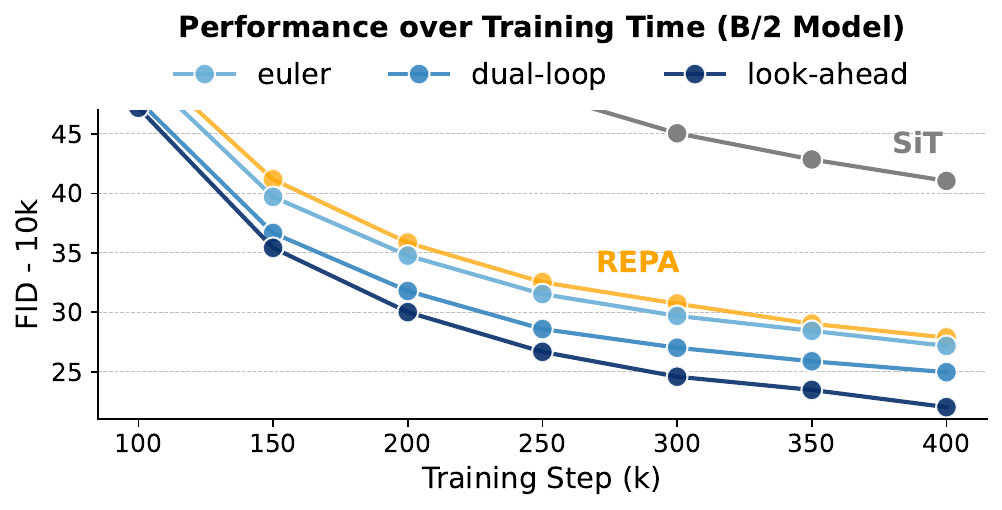}
    \caption{\textbf{Representation Alignment} comparison on B/2 models. Patch Forcing (blue lines) is orthogonal to {\color{repa}REPA}~\cite{yurepresentation_repa} and we can observe similar training improvements compared to standard {\color{sit}SiT} training~\cite{ma2024sit}. Integrating REPA into our PFT model yields further gains, while our {\color{dualloop} \textit{dual-loop}} and {\color{lookahead}\textit{look-ahead}} samplers both improve upon baseline {\color{euler}Euler} sampling with our REPA-PF model. We keep all NFEs fixed to ensure a fair comparison.
    }
    \label{fig:supp:repa-over-time}
\end{figure}

\paragraph{Orthogonality to REPA}
\Cref{fig:supp:repa-over-time} shows performance over training iterations when integrating REPA into PF and compares it to SiT~\cite{ma2024sit} and REPA~\cite{yurepresentation_repa}. Our PFT consistently improves over REPA, yielding additive gains already under standard Euler sampling. Applying our uncertainty-aware samplers on top provides further improvements. For a fair comparison, we keep the number of function evaluations (NFE) fixed across all sampling strategies.

\begin{figure}
    \centering
    \includegraphics[width=0.8\linewidth]{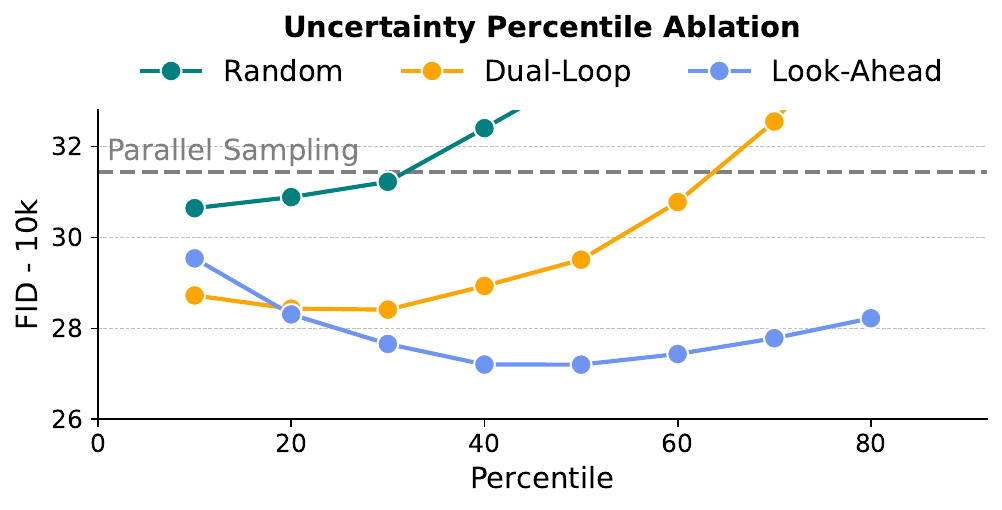}
    \caption{\textbf{Ablation of patch-difficulty threshold} with Patch Forcing B/2 model. We ablate over the percentile for confident pixels at sampling. Our samplers perform best at around the 40\% percentile and outperform the parallel sampling baseline.
    }
    \label{fig:supp:abl-percentiles}
\end{figure}

\paragraph{Difficulty Percentile at Sampling}
We ablate the confidence threshold used to select patches for early progression in the PFT-B/2 model across different sampling strategies in \cref{fig:supp:abl-percentiles}. Specifically, we vary the percentile cutoff that determines which patches are considered “confident” and thus used to provide context for others.
Interestingly, random selection can offer slight improvements over the parallel sampling baseline, suggesting that even naive context can help. However, the benefits are substantially greater when confident patches are selected based on predicted uncertainty. Both of our proposed samplers: \textit{dual-loop} and \textit{look-ahead} samplers, consistently outperform random sampling and parallel sampling, confirming the value of informed, uncertainty-driven patch scheduling.

\begin{algorithm}[t]
\footnotesize
\caption{Look-ahead Sampling}
\label{alg:fm-context-transport}
\begin{algorithmic}[1]
\Require Model $f_\theta(x,t)\to (v,\; \mathbf{uc})$ where $v$ is velocity and $\mathbf{uc}$ is uncertainty map, $x\in \mathbb{R}^{B\times C\times H\times W}, t\in \mathbb{R}^{B\times H\times W}$;
uncertainty percentile $p \in (0,1)$; 
context factor $\alpha>1$; 
time grid $0=t_0<t_1<\dots<t_K=1$; 
Euler stepper $\mathrm{Step}(x,t,t'):=x+(t'-t)\,v$ (\emph{replace with other ODE solver if desired})
\State Initialize $x_0 \sim \mathcal{N}(0,I)$ \Comment{start from noise at $t_0=0$}
\For{$k=0$ to $K-1$}
  \State $x \gets x_k$, \quad $t \gets t_k$, \quad $t_{\text{next}} \gets t_{k+1}$
  \State $(v_t,\; \mathbf{uc}_t) \gets f_\theta(x, t)$ \Comment{predict velocity and uncertainty}
  \State $\tau_p \gets \mathrm{Percentile}(u_c,\; p)$ \Comment{adaptive thresholding}
  \State $M_{\text{conf}} \gets \mathbf{1}[\,\mathbf{uc} \le \tau_p\,]$; \  $M_{\text{unc}} \gets 1 - M_{\text{conf}}$ \Comment{confidence masks}
  \State $t_\text{ctx} \gets \min(\alpha\, t,\; 1)$ %
  \State $x_{\text{ctx}} \gets \mathrm{Step}(x,\; t,\; t_\text{ctx})$ \Comment{one-step look-ahead}
  \State $\tilde{x} \gets M_{\text{conf}}\odot x_{\text{ctx}} \;+\; M_{\text{unc}}\odot x$ \Comment{use look-ahead for $M_{\text{conf}}$}
  \State $\tilde{t} \gets M_{\text{conf}}\odot t_\text{ctx} \;+\; M_{\text{unc}}\odot t$ \Comment{use look-ahead for $M_{\text{conf}}$}
  \State $(v_{\text{ctx}},\; \underline{\hspace{0.3cm}}\ ) \gets f_\theta(\tilde{x},\, \tilde{t})$ \Comment{context-aware velocity}
  \State $v_{\text{final}} \gets M_{\text{unc}}\odot v_{\text{ctx}} \;+\; M_{\text{conf}}\odot v$ \Comment{replace $M_\text{unc}$ prediction}
  \State $x \gets x + (t_{\text{next}}-t)\, v_{\text{final}}$ \Comment{advance to $t_{k+1}$}
\EndFor
\State \Return $x$ %
\end{algorithmic}
\end{algorithm}

\begin{figure}
    \centering
    \includegraphics[width=0.8\linewidth]{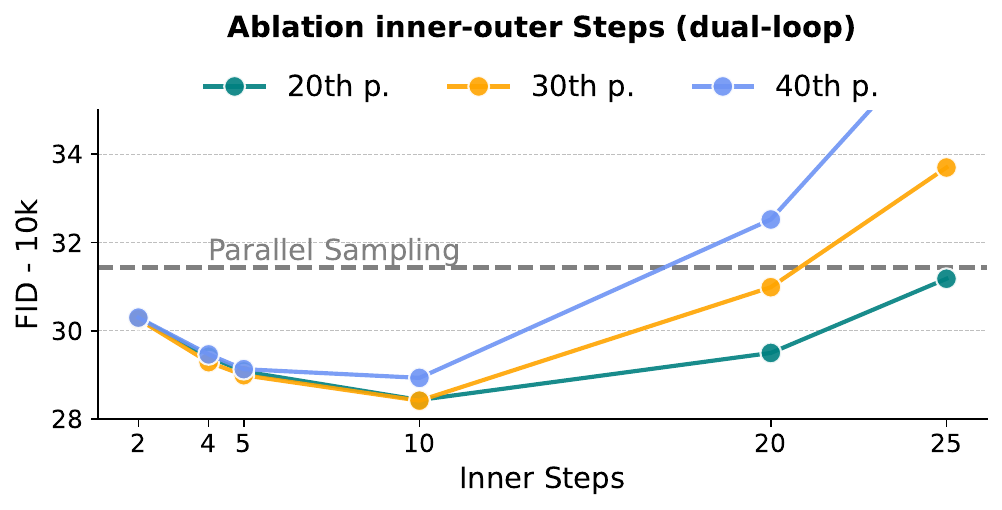}
    \caption{\textbf{Ablation of dual-loop hyperparameters} with Patch Forcing B/2 model. We ablate over percentiles of confident patches and also vary the number of inner vs outer steps while keeping the total NFE at $\text{inner}\times\text{outer}=100$. The model achieves optimal performance when the inner and outer steps are balanced.
    }
    \label{fig:supp:abl-inner-outer}
\end{figure}

\paragraph{Inner vs. Outer Steps}
In the dual-loop sampler, we can freely choose the number of inner vs outer steps, as well as the percentile of confident patches. We conduct ablations under a fixed $\text{NFE}=100$ in \Cref{fig:supp:abl-inner-outer}. We vary the number of inner and outer loop steps to identify the optimal configuration, and ablate over different confidence percentiles used to select the patches for context propagation.
We find that using 10 inner steps and 10 outer steps yields the best performance. When the number of outer steps is too high, the behavior begins to resemble standard Euler sampling, diminishing gains from the uncertainty-aware sampling. Conversely, too few outer steps lead to overly aggressive updates, resulting in inaccurate context and degraded performance. This effect is particularly pronounced when a larger percentage ($40\%$) of confident pixels are advanced with insufficient outer steps, as reflected by a sharp drop in generation quality.

\paragraph{Classifier-free Guidance}
We show that our method also works well with standard classifier-free guidance (CFG) \cite{ho2021classifier_cfg}. In \cref{fig:supp:cfg-ablation} we show, similar to previous findings, increasing the CFG scale leads to visually better results, but at the cost of reduced diversity.

\paragraph{Connection to Region-Adaptive Sampling}
As discussed in related works, RAS~\cite{liu2025region} is fundamentally an inference-time acceleration method. It focuses on \textit{computational efficiency during inference via caching} and reusing patch computations across timesteps with
\textit{minor fidelity degradation} at fixed NFEs.
In contrast, PF introduces a training paradigm: we train the model with patch-wise timesteps, enabling it to reason over heterogeneous noise states, which already improves generation quality. Building on this, we design samplers that improve image fidelity further by propagating confident patches to provide context for harder ones. These components are designed for performance gains rather than acceleration. Thus, our method and RAS are \textit{orthogonal and can be combined}: \cref{tab:compute_tradeoff} shows that our method integrates with previous-step prediction caching and KV caching from RAS to achieve a favorable compute-performance trade-off, even with a simple integration of our dual-loop sampler (last column).

\begin{table}[t]
\scriptsize
\centering
\setlength{\tabcolsep}{3pt}
\begin{tabular}{l @{\hspace{20pt}} c c @{\hspace{20pt}} c c c}
\toprule
 & SiT-B & {\tiny +RAS} & PF-B & {\tiny +Dual-loop} & {\tiny +Dual-loop,Cache} \\
{\scriptsize Sampling Ratio (\%)} & 100 & 60 & 100 & 100 & 60 \\
\lightmidrule
{\scriptsize NFE=$16$}  & 48.0 & 50.9 & 43.9 & 42.5 & 45.0 \\
{\scriptsize NFE=$32$}  & 43.1 & 45.2 & 33.9 & 32.4 & 36.3 \\
{\scriptsize NFE=$250$} & 41.0 & 41.2 & 31.1 & 28.2 & 28.5 \\
\bottomrule
\end{tabular}
\caption{
\textbf{ImageNet FID$_\text{10K}$ compute-fidelity tradeoff.}
PF + caching outperforms RAS with identical model and compute.
}
\label{tab:compute_tradeoff}
\vspace{-3mm}
\end{table}

\subsection{Text-conditional Generation}
\label{sec:supp:t2i}

\begin{figure*}[t]
\centering

\setlength{\tabcolsep}{4pt} %
\newcommand{\rowy}[2]{
\vspace{-2.6cm}
\footnotesize \textit{#2} &
\includegraphics[width=0.42\linewidth]{fig/t2i/text-rendering/#1-base.jpg} &
\includegraphics[width=0.42\linewidth]{fig/t2i/text-rendering/#1-pft.jpg} \\
}
\begin{tabular}{p{0.13\linewidth} c c}
 & \textbf{Flow Matching Baseline} & \textbf{Patch Forcing} \\
\midrule 
\rowy{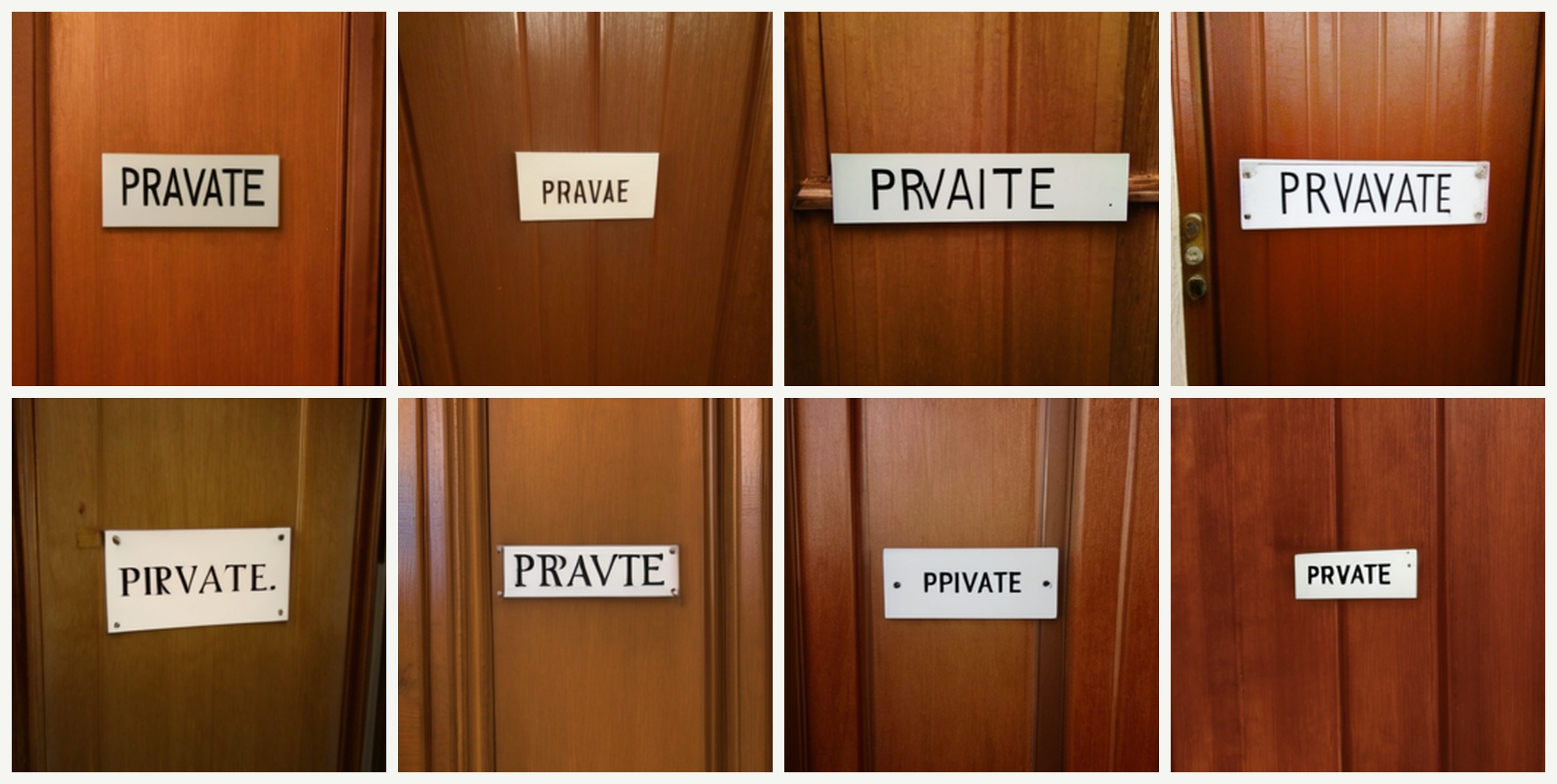}{A simple door sign that reads PRIVATE mounted on a wooden door.}
\rowy{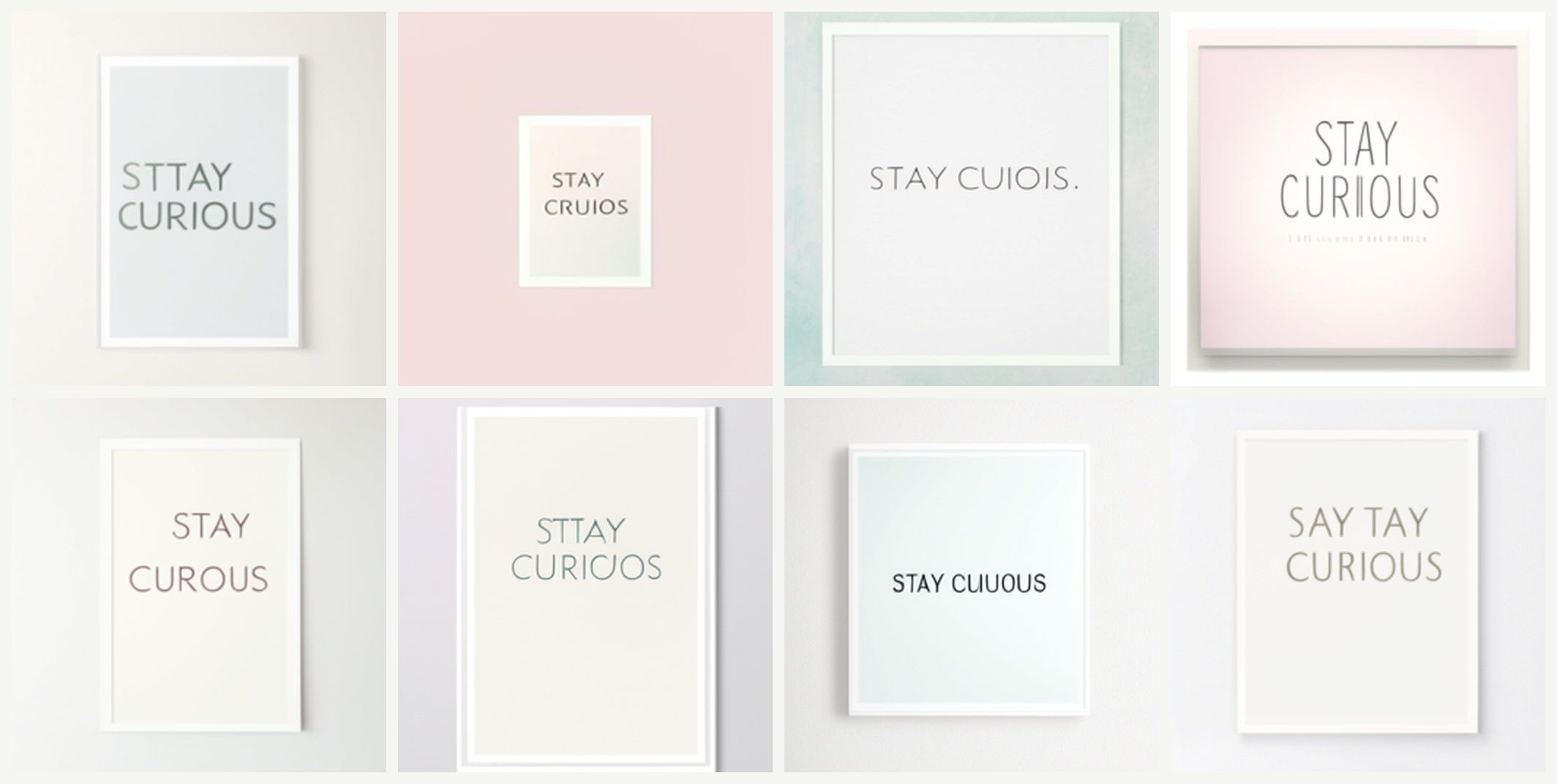}{A minimal poster with large centered text STAY CURIOUS. Clean layout soft colors.}
\rowy{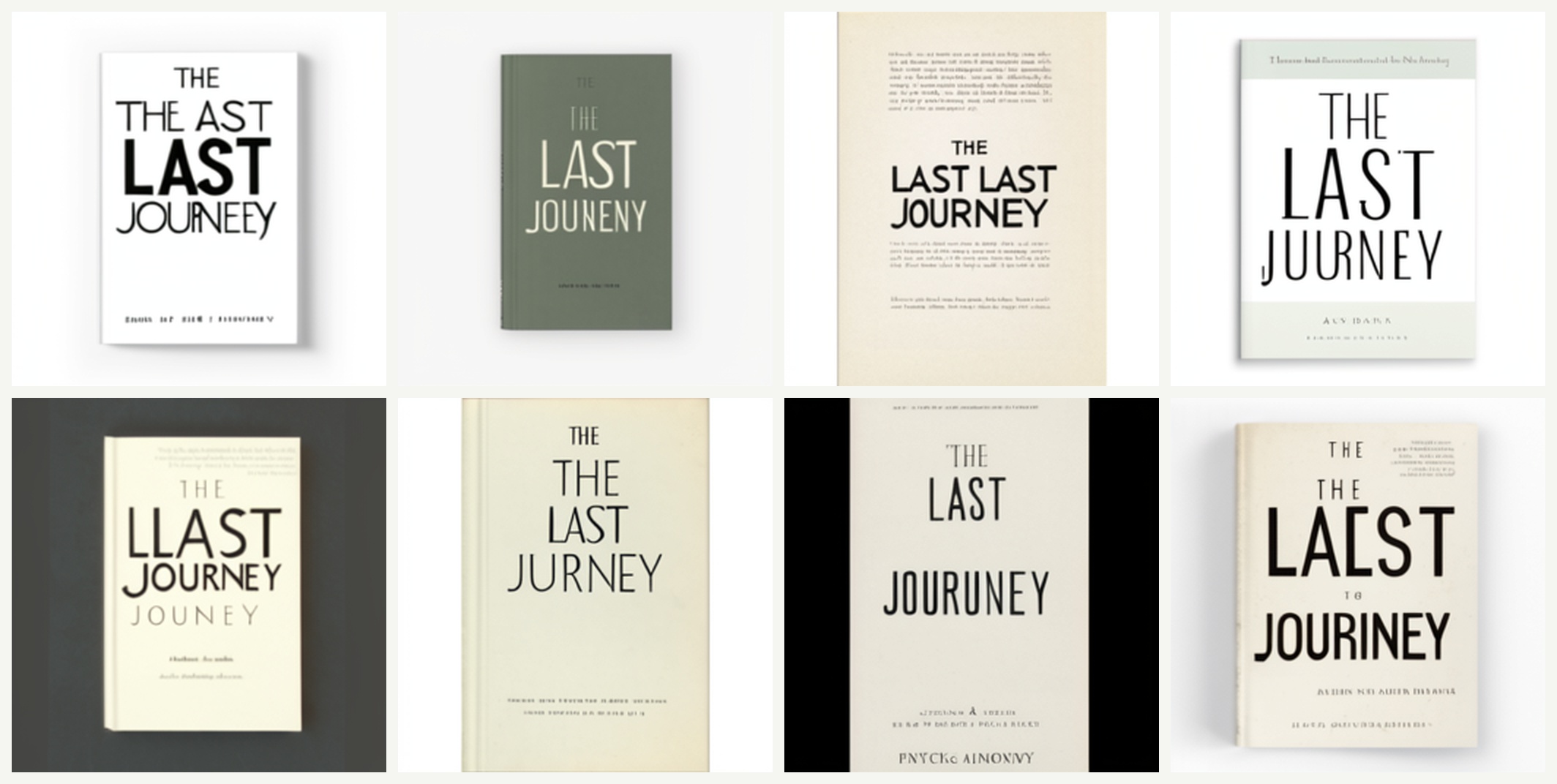}{A book cover with bold title text THE LAST JOURNEY. Minimal design.}
\rowy{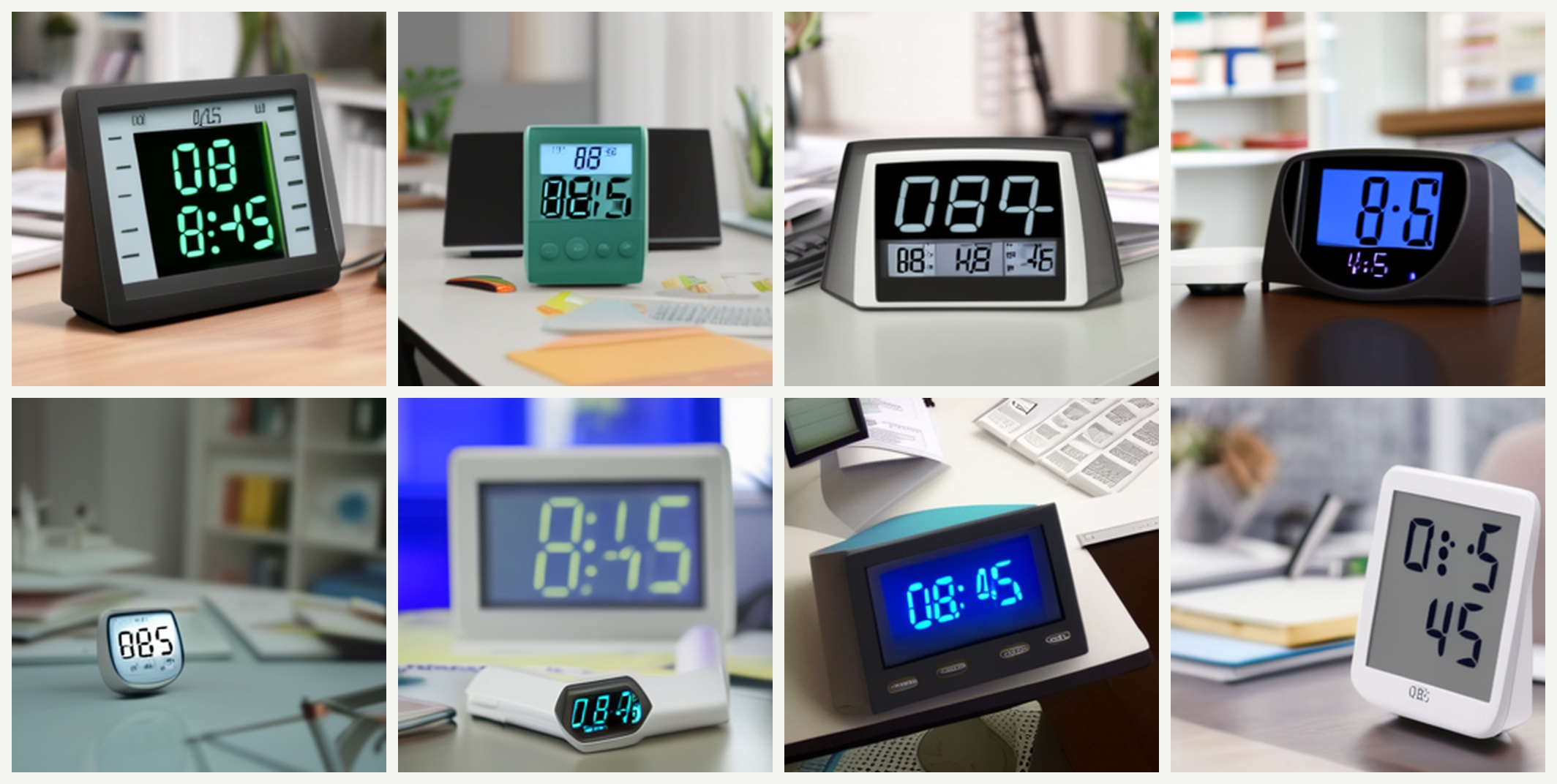}{A digital clock displaying 08:45 on a study table.}
\rowy{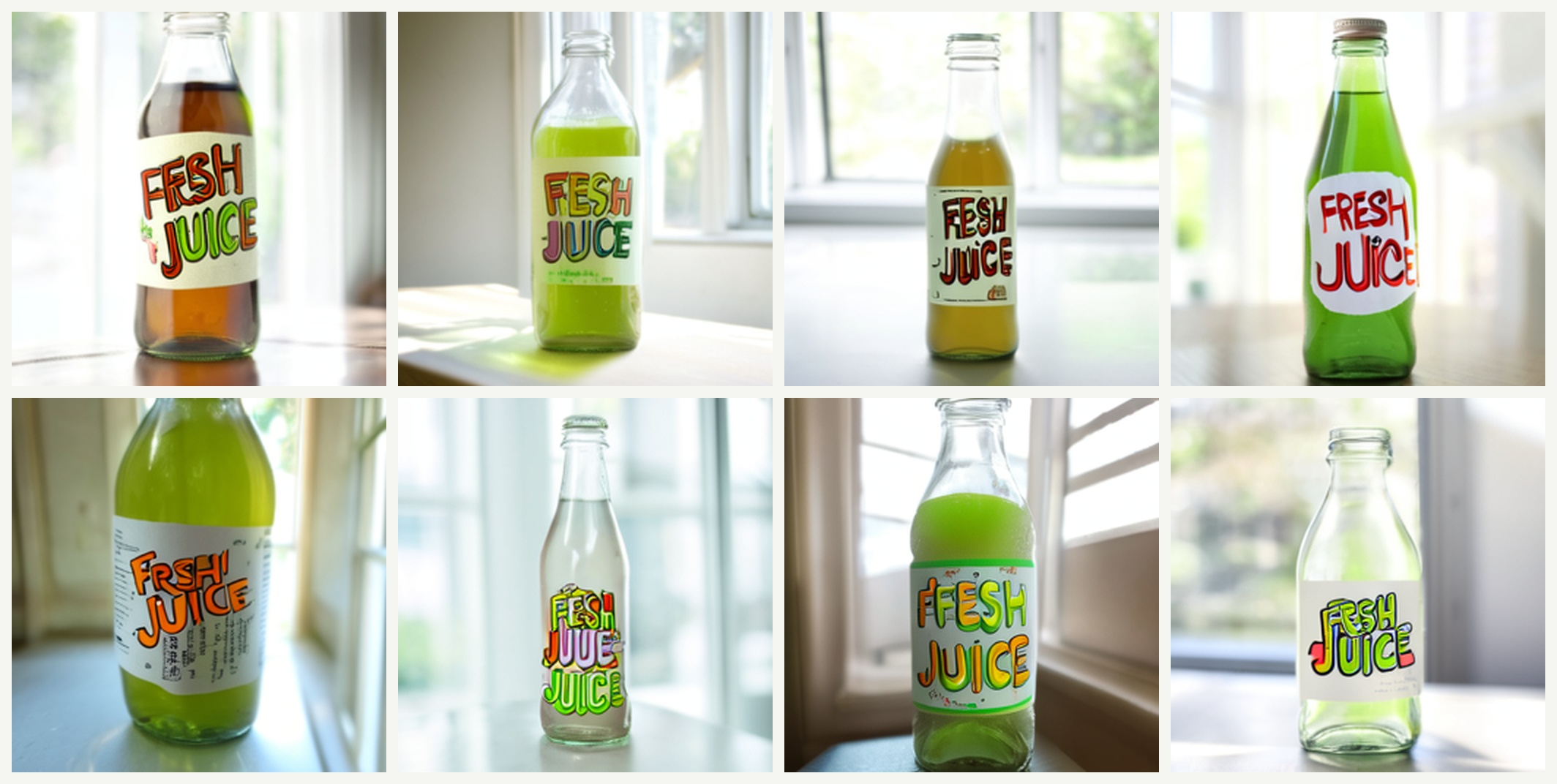}{A glass bottle with big label text FRESH JUICE in cartoon font. Natural light.}
\end{tabular}
\caption{\textbf{Uncurated Text-to-Image $256$ px samples} from the baseline Flow Matching model with Logit-Normal schedule compared to our PFT model. We keep the model architecture, data, training, and sampling fixed (same amount of Euler steps and seed) and only change the training paradigm from plain Flow Matching to Patch Forcing.}
\label{fig:supp:qualitative-text-rendering}
\end{figure*}

\paragraph{Text Rendering Quality.}
\Cref{fig:supp:qualitative-text-rendering} presents uncurated samples from two text-to-image models: one trained with standard Flow Matching (FM) and one trained with Patch Forcing. Following \cite{esser2024scaling-sdv3}, we use logit-normal timestep sampling for the FM baseline, while the Patch Forcing Transformer (PFT) is trained with our proposed Logit-Normal Truncated Gaussian (LTG) sampler. Both models share exactly the same architecture, text encoder stack, and training data; the only difference is the transition from Flow Matching training to Patch Forcing training with heterogeneous patch-wise noise levels. Overall, we observe that our PFT produces clearer text compared to the FM baseline.

We further evaluate this effect quantitatively by measuring text rendering accuracy using an OCR-based evaluation protocol. Specifically, we construct a set of prompts that explicitly require rendering text (e.g., "A man holding a sign that reads '…'") and generate images from both models. We then apply an off-the-shelf OCR model (EasyOCR) to extract the rendered text and compare it to the ground-truth prompt text. We generate the prompt set based on $10$ template texts and $10$ inner texts, and sample $8$ images per prompt, resulting in $800$ evaluation images.
\Cref{tab:supp:ocr} shows that our PFT consistently improves text rendering quality over the Flow Matching baseline. The qualitative improvements observed in \Cref{fig:supp:qualitative-text-rendering} are also reflected quantitatively by higher exact match rates and lower Levenshtein distances. In particular, PFT with Euler and Dual Loop sampling achieves substantial gains across all metrics. Interestingly, the Look-Ahead sampler, while often performing favorably on standard image quality metrics, shows the weakest text rendering performance among the PFT variants. A possible explanation is that, in the Look-Ahead setting, the context is fixed during generation, limiting the model’s ability to iteratively refine and correct text. In contrast, Dual Loop sampling allows for limited inner updates, which might explain why its performance remains closer to the Euler baseline.

\begin{table}[t]
\centering
\footnotesize
\resizebox{\linewidth}{!}{
\begin{tabular}{lccc}
\toprule
Model & Exact Match Rate $\uparrow$ & Mean Lev. $\downarrow$ & Mean Norm. Lev. $\downarrow$ \\
\midrule
Flow Matching & 0.3937 & 9.5400 & 0.3348 \\
PFT + Euler & 0.6162 & 5.5462 & 0.2221 \\
PFT + Dual Loop & 0.6125 & 5.8300 & 0.2226 \\
PFT + Look Ahead & 0.4875 & 4.7313 & 0.2911 \\
\bottomrule
\end{tabular}
}
\caption{
\textbf{OCR-based text rendering comparison} for a Flow Matching Baseline with our Patch Forcing Transformer. Both models share the same architecture, text encoder stack, number of parameters, data, and number of function evaluations.
}
\label{tab:supp:ocr}
\end{table}

\begin{figure}
    \centering
    \footnotesize
    \setlength\tabcolsep{0.5pt}
    \newcommand{\imagepng}[1]{
        \includegraphics[width=0.18\linewidth]{fig/supp/t2i-adapt/#1}
    }
    \begin{subfigure}{\linewidth}
    \centering
    \begin{tabular}{ccccc}
    & \multicolumn{4}{c}{
        $\xrightarrow{\hspace{2cm}}\;
        \text{less noise}\;
        \xrightarrow{\hspace{2cm}}$
    }
    \\
    \rotatebox{90}{\hspace{1.8em}$\hat{x}_{t}$} &
        \imagepng{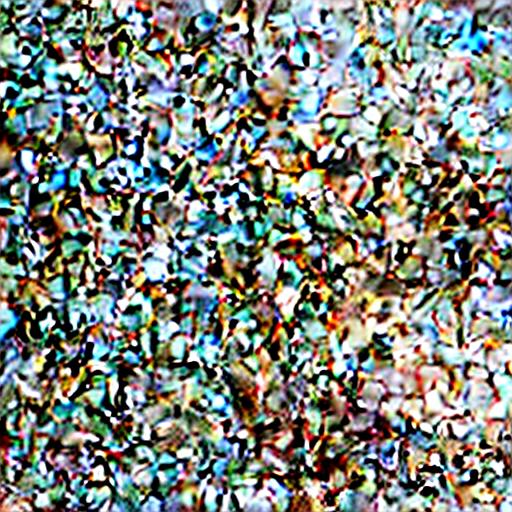} &
        \imagepng{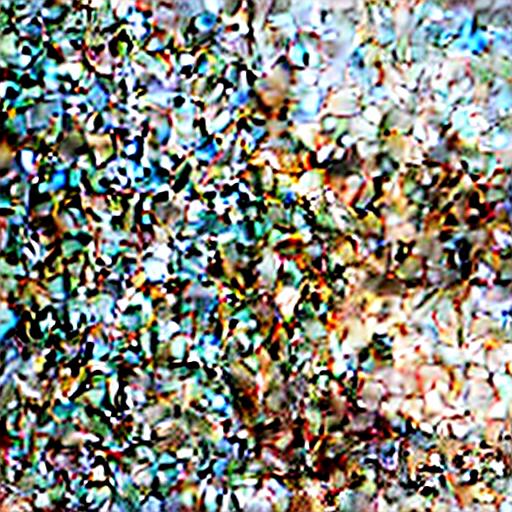} &
        \imagepng{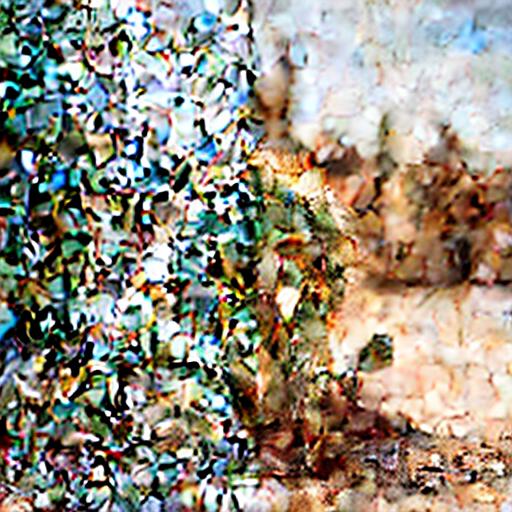} &
        \imagepng{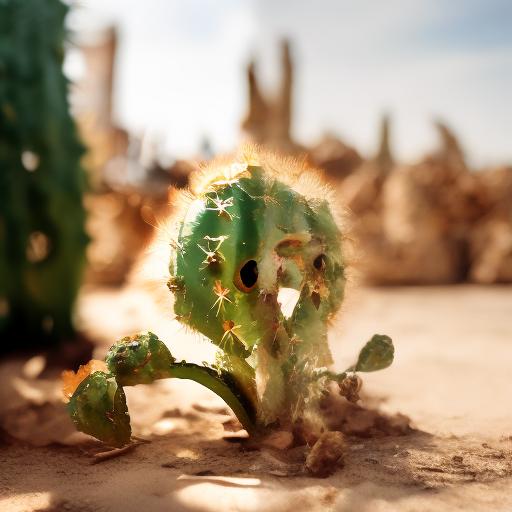}
        \\
    \rotatebox{90}{\hspace{1.2em}$\hat{x}_{t \rightarrow 1}$} &
        \imagepng{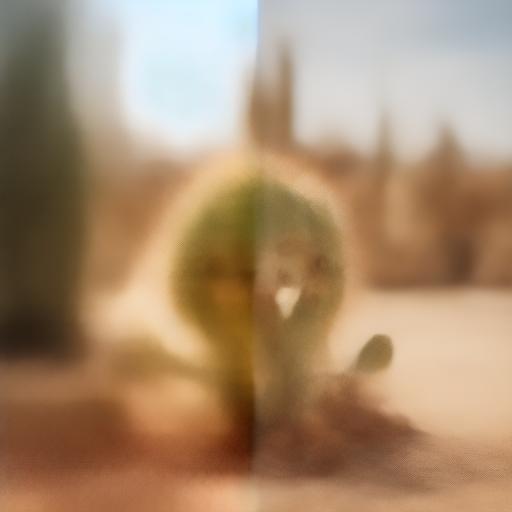} &
        \imagepng{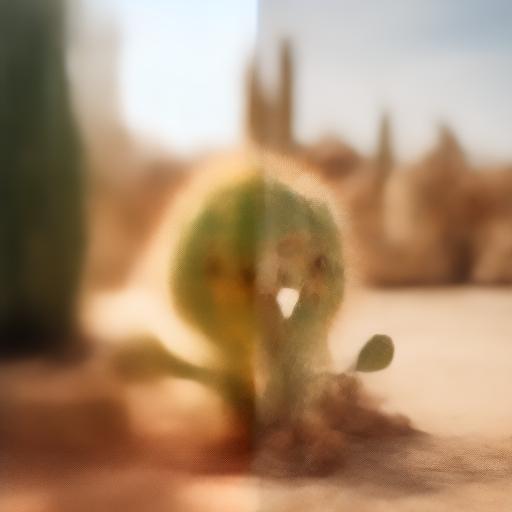} &
        \imagepng{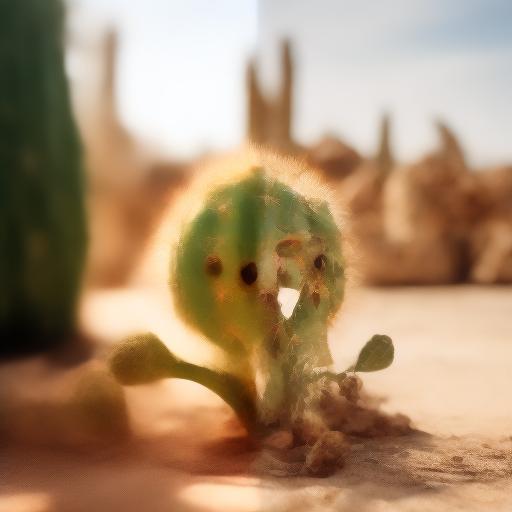} &
        \imagepng{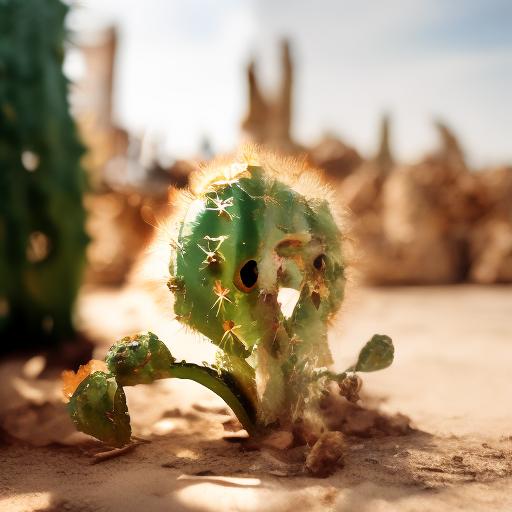}
        \\
    \rotatebox{90}{Uncertainty} &
        \imagepng{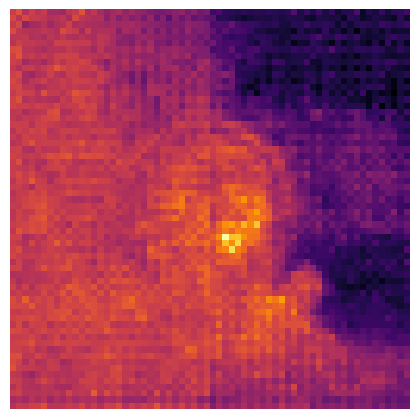} &
        \imagepng{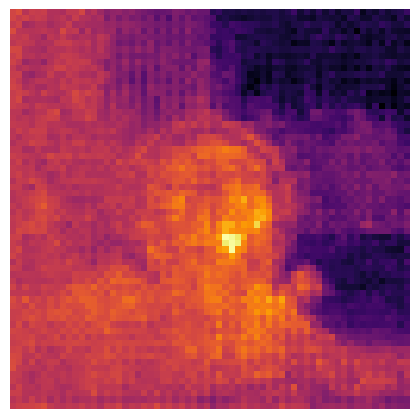} &
        \imagepng{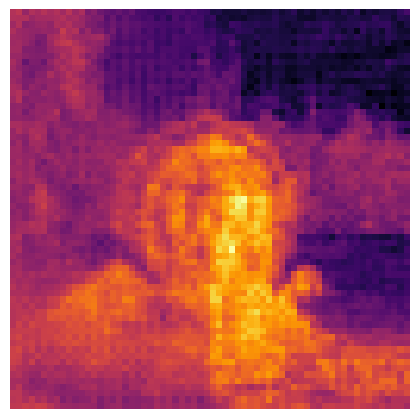} &
        \imagepng{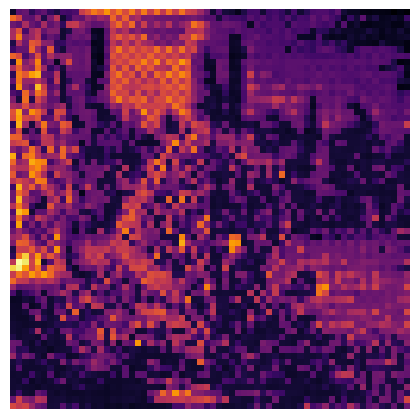}
        \\
      \vspace{0.1cm}
    \end{tabular}
    \end{subfigure}
    \caption{
    \textbf{PixArt-$\alpha$ with heterogeneous denoising.}
    We adapt a pretrained text-to-image model (PixArt-$\alpha$~\cite{chen2023pixartalpha}) and observe that it can handle per-patch noise levels to some degree. The $\sigma$ prediction of it can be repurposed for our samplers. We do not use CFG in this example.
    }
    \label{fig:supp:adapting-t2i}
\end{figure}

\paragraph{Adapting a Pre-Trained T2I Model.}
We additionally explore whether our samplers can be applied \emph{zero-shot} to an existing pretrained text-to-image model.
Since our method requires a patch-difficulty prediction, PixArt-$\alpha$~\cite{chen2023pixartalpha} is one of the few suitable candidates, as it inherits a $\sigma$-prediction head from the original Diffusion Transformer~\cite{peebles2023scalable-dit}. Hence, we repurpose the model’s variance prediction as a proxy for patch difficulty by averaging it across channels, which we find to align with difficult regions (see \Cref{fig:supp:adapting-t2i}).
Although PixArt-$\alpha$ is not trained with patch-wise timesteps, we find that it is relatively robust to spatially varying noise scales (see \Cref{fig:supp:adapting-t2i}). Therefore, to apply our sampling strategy, we broadcast the timestep conditioning to the spatial token level such that each latent patch can be assigned its own noise level. Taken together, this enables uncertainty-aware sampling without any weight modification or fine-tuning.
\cref{tab:supp:t2i-compbench-pixart} shows that applying our samplers zero-shot to PixArt-$\alpha$ improves image quality over the standard Euler sampler on T2I-CompBench++~\cite{huang2023t2i}. Although this analysis is exploratory, it suggests that diffusion transformers can tolerate spatially varying noise levels even when trained only with homogeneous timesteps. Patch Forcing amplifies this effect by explicitly training with patch-wise noise scales and thereby reducing the train–test gap, which makes our samplers more effective.

\begin{table}[t]
\centering
\resizebox{\linewidth}{!}{
\begin{tabular}{lcccccc}
\toprule
& \makecell{Color \\ B-VQA}
& \makecell{Shape \\ B-VQA}
& \makecell{Texture \\ B-VQA}
& \makecell{2D-Spatial \\ UniDet}
& \makecell{3D-Spatial \\ UniDet}
& \makecell{Non-Spatial \\ CLIP} \\
\midrule
Euler            & 0.3186 & 0.3264 & 0.3394 & 0.0530 & 0.1793 & 0.2793 \\
Dual-loop        & \textbf{0.3285} & \textbf{0.3371} & \textbf{0.3448} & 0.0608 & 0.1753 & 0.2782 \\
Look-ahead       & 0.3126 & 0.3313 & 0.3348 & \textbf{0.0625} & \textbf{0.1925} & \textbf{0.2814} \\
\bottomrule
\end{tabular}
}
\caption{\textbf{Evaluation of sampling strategies on PixArt-$\alpha$.} We evaluate sampling strategies \textbf{zero-shot} on the pre-trained PixArt-$\alpha$ model~\cite{chen2023pixartalpha} using the T2I-CompBench++ benchmark~\cite{huang2023t2i}, where the model is not trained with varying patch-wise timesteps. All experiments are evaluated without CFG.
}
\label{tab:supp:t2i-compbench-pixart}
\end{table}

\paragraph{Comparison to Self-Flow}
Concurrent to our work, \textit{Self-Flow}~\cite{chefer2026selfflow} also introduces heterogeneous noise levels during diffusion training. Similar to our findings, the authors observe that naively applying diffusion forcing at the patch/token level substantially degrades generation quality. To address this issue, Self-Flow proposes Dual-Timestep sampling, which uses only two distinct noise scales during training.
While this reduces the train-test gap, we approach the issue from a different perspective. Sampling the timesteps per patch from a uniform distribution results in an average noise level around $t \approx 0.5$ during training, whereas at inference, the model must start from full noise, with no prior information available. To close this gap, we directly control the maximum information during training via our LTG sampler. Furthermore, while Self-Flow primarily leverages heterogeneous noise levels for representation learning, we additionally show that they can be exploited for non-uniform denoising through our proposed sampling strategies.

Similar to Self-Flow, we also observe improved text rendering for our T2I model trained with Patch Forcing (\Cref{sec:supp:t2i}). This suggests that the gains may stem from heterogeneous noise scales during training rather than the specific representation loss used in Self-Flow, which may render the additional teacher–student forward passes unnecessary. Understanding what exactly drives these improvements is an interesting direction for future work.

\section{Implementation Details}

We detail all implementation details in the following sections for ImageNet and our text-to-image experiments.

\subsection{Class-conditional ImageNet}
We follow the training setup of~\cite{ma2024sit}, using a batch size of $256$ and AdamW~\cite{loshchilov2017adamw} with a constant learning rate of $1\times10^{-4}$. We maintain an exponential moving average (EMA) of model parameters throughout training with a decay of $0.9999$, and report results using the EMA weights. The only data augmentation applied is random horizontal flipping with probability $0.5$. For metrics, we follow the ADM evaluation suite \cite{dhariwal2021diffusion}. Results reported in \cref{tab:dit-cfg} are from our PF-XL model with representation alignment~\cite{yurepresentation_repa}.

The original SiT implementation~\cite{ma2024sit} requires only minimal changes to support patch-wise noise levels. In \cref{fig:supp:code-sit} we show the modifications to the Diffusion Transformer. With these adjustments, the only remaining change is to adapt the timestep sampler used during training.

\paragraph{LTG Scheduler Implementation Details}
As discussed in the paper, we design a truncated Gaussian sampler that samples only from the lower (noisier) half of the Gaussian distribution. However, since timesteps must remain within the valid range $[0, 1]$, we cannot directly apply arbitrary combinations of $\text{t}_{\max}$ and standard deviation std without risk of sampling invalid values.
To ensure the sampled timesteps remain within a meaningful range, we dynamically adjust the standard deviation based on $\text{t}_{\max}$. Specifically, we set
$\text{std}_{\text{eff}} = \min(\text{t}_{\max}/2, \text{std})$,
so that, according to the empirical 2-standard-deviation rule, approximately $95\%$ of the mass of the Gaussian distribution lies above $0$. For rare cases where sampled values fall below 0, we replace them with random values uniformly sampled from $[0, t_\text{max}]$.

\begin{figure*}[tb]
    \centering
    \includegraphics[width=0.98\linewidth]{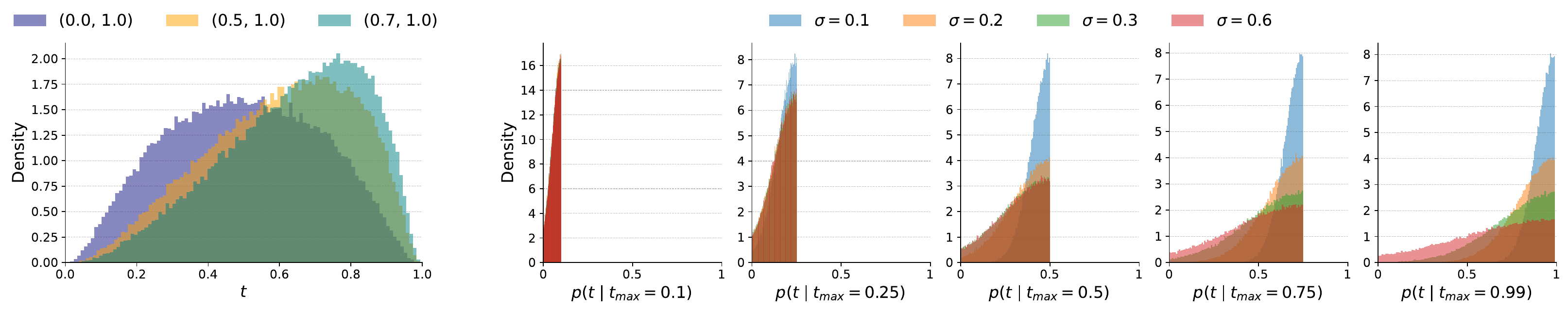}
    \caption{\textbf{Logit-Normal Truncated Gaussian Timestep Schedule} \textit{Left}: We first sample $t_{max}$ from a Logit-Normal distribution with location and scale parameters $(m, s)$ according to \cite{esser2024scaling-sdv3}. \textit{Right}: Given $t_{max}$, we sample the individual patch timesteps according to a truncated Gaussian with parameter $\sigma$.
    }
    \label{fig:supp:ltg-explain}
\end{figure*}

For our LTG sampler we have three parameters. First, the location $m$ and scale $s$ parameters for the Logit-Normal $t_{\max}$ sampling (see \Cref{fig:supp:ltg-explain} left) and second, the $\sigma$ parameter controlling the spread of the lower half around the sampled $t_{\max}$. We provide the pseudo-code in \Cref{alg:ltg-sampler}.

\begin{algorithm}[t]
\caption{LTG Sampler Pseudo-code}
\label{alg:ltg-sampler}

\begin{lstlisting}
# loc and scale: logit-normal params for t_max
# std: Gaussian spread for t

t_max = lognorm(loc + scale * randn(bs))
std = min(t_max / 2, std)

t_max = t_max[:, None]
std = std[:, None]
eps = randn(bs, dim)

t = t_max - abs(eps) * std
t[t < 0] = rand_like(t) * t_max # reset negatives

return t
\end{lstlisting}
\end{algorithm}

\begin{figure}
    \centering
    \includegraphics[width=0.7\linewidth]{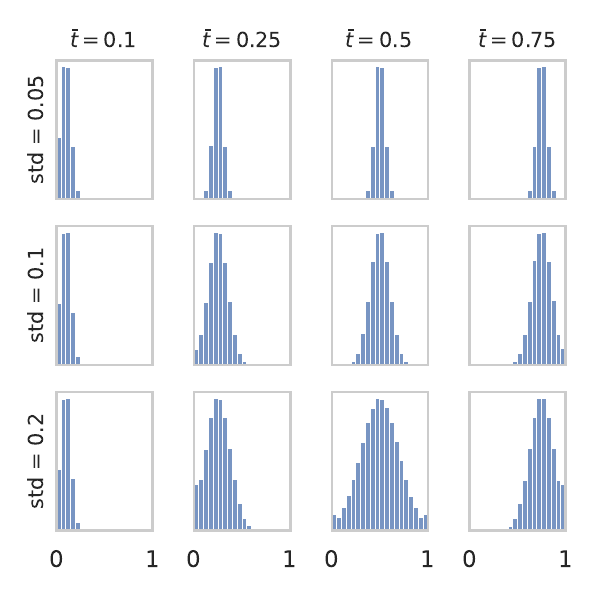}
    \includegraphics[width=0.7\linewidth]{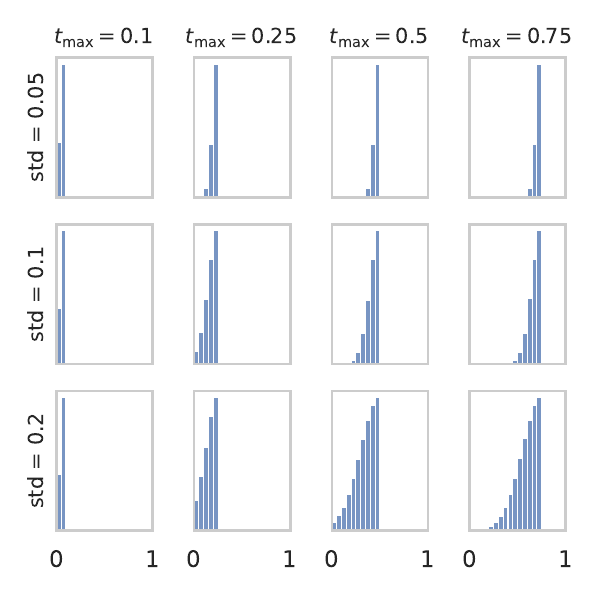}
    \caption{\textbf{Visualization of our timesteps samplers}: the first is a Gaussian distribution-based sampler, while the second is a truncated Gaussian sampler.}
    \label{fig:placeholder}
\end{figure}

\subsection{Text-to-Image}
For our text-to-image experiments, we train a 1.2B Patch Forcing Transformer (PFT) on a 120M subset of COYO~\cite{kakaobrain2022coyo-700m}. We first recaption all images with InternVL3-2B~\cite{zhu2025internvl3} using long-form descriptions, and then distill this long caption into three variants: \emph{long}, \emph{medium}, and \emph{keyword} captions. During training, we sample uniformly from these caption variants, and with probability $0.1$, we replace the caption with an empty prompt to enable classifier-free guidance~\cite{ho2021classifier_cfg}. We encode text with Qwen3-1.7B~\cite{qwen3vlembedding}, and following~\cite{ma2024exploring}, we insert a lightweight two-layer text refiner transformer of width $1536$ between the frozen text features and the cross-attention blocks of our diffusion transformer. For all T2I models, we use the FLUX.2 autoencoder~\cite{flux-2-2025}.

\begin{figure}[t]
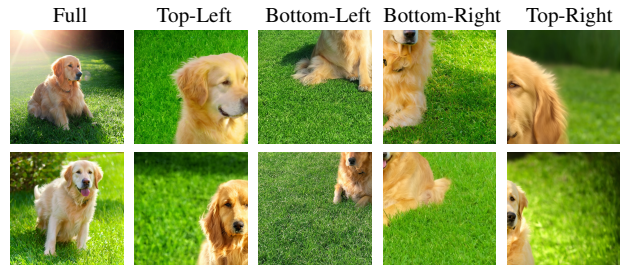

    \centering \small
    \setlength\tabcolsep{1pt}
    \newcommand{\baseimg}[2]{
    \includegraphics[width=0.18\linewidth]{fig/t2i/crop-cond/#1-0000#2.png}
    }
    \newcommand{\rowy}[5]{
    \baseimg{full}{#1} &
    \baseimg{top-left}{#2} &
    \baseimg{bottom-left}{#3} &
    \baseimg{bottom-right}{#4} &
    \baseimg{top-right}{#5}
    }
    \begin{tabular}{ccccc}
        \footnotesize Full & \footnotesize Top-Left & \footnotesize Bottom-Left & \footnotesize Bottom-Right & \footnotesize Top-Right \\
        \rowy{7}{4}{4}{2}{4} \\
        \rowy{3}{8}{7}{4}{1} \\
    \end{tabular}
    \caption{
    \textbf{Crop-size conditioning} via RoPE.
    }
    \label{fig:supp:crop-size-cond}
\end{figure}

Similar to prior work, such as SDXL~\cite{podellsdxl}, we employ crop-size conditioning during training. Since our PFT uses RoPE~\cite{su2021roformer}, we directly integrate crop-size conditioning via positional encoding, adapting relative positions to the sampled crop. At inference time, we always set the crop size to the full image resolution. \Cref{fig:supp:crop-size-cond} shows how this conditioning mechanism can be used to generate different crop-outs for the same prompt.

We pre-train PFT for 400k iterations at $ 256$ px resolution with a batch size of $1024$ and a fixed learning rate of $10^{-4}$, and then finetune it for an additional 50k iterations at $512$ px resolution on a mixture of high-aesthetics filtered COYO and JourneyDB~\cite{sun2023journeydb}. For all models, we maintain an EMA with a decay factor of $0.9999$ and report results using the EMA weights. We set the uncertainty loss weight to $0.01$ for all experiments, following~\cite{wewer2025spatial_srm}.

For the comparison in \Cref{fig:supp:qualitative-text-rendering,fig:t2i-spelling}, we additionally train a vanilla Flow Matching baseline on exactly the same data, with the same architecture, batch size, learning rate, and training schedule. The only difference lies in the training timestep sampling strategy: the vanilla model uses the logit-normal schedule from~\cite{esser2024scaling-sdv3} and broadcasts a single timestep to all spatial tokens, whereas the PFT samples different timesteps per patch using our Logit-Normal Truncated Gaussian sampler.

\begin{figure}[htbp]
  \centering
  \setlength{\tabcolsep}{2pt}
  \begin{tabular}{cccc}
    \includegraphics[width=0.25\linewidth]{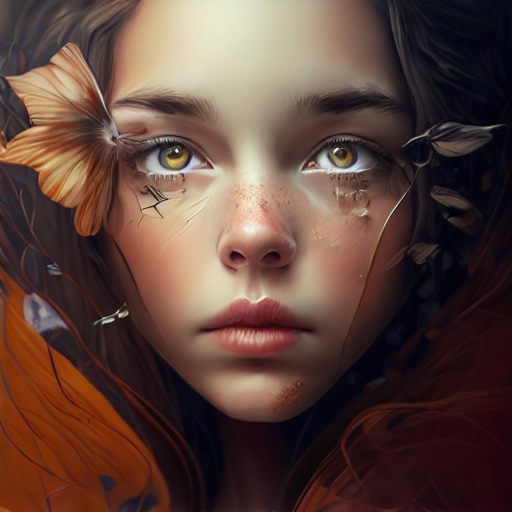} &
    \includegraphics[width=0.25\linewidth]{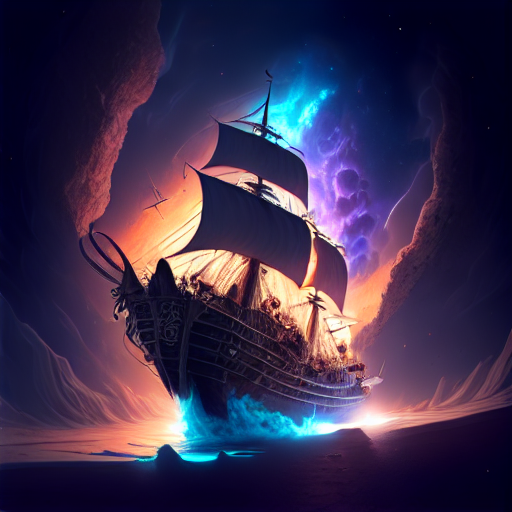} &
    \includegraphics[width=0.25\linewidth]{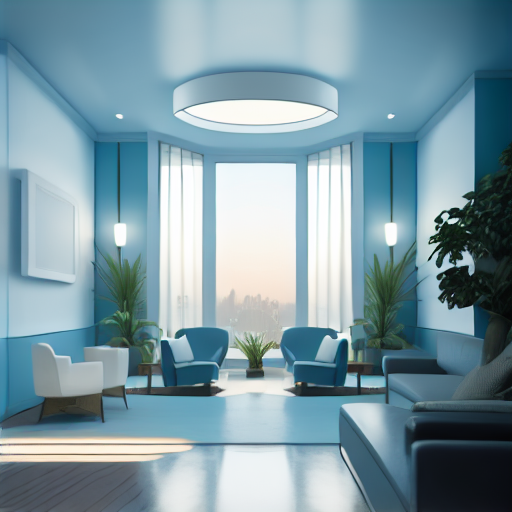} &
    \includegraphics[width=0.25\linewidth]{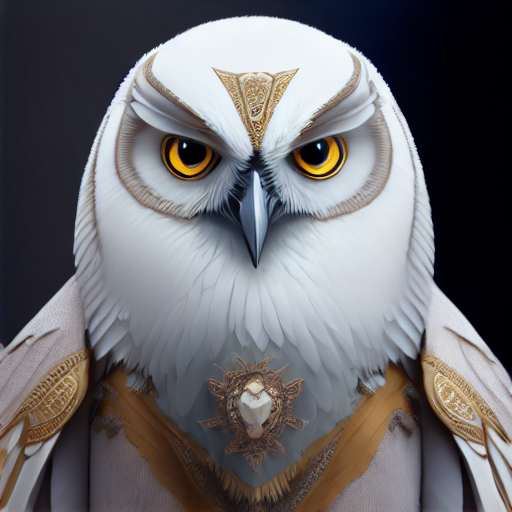}
    \\
    \includegraphics[width=0.25\linewidth]{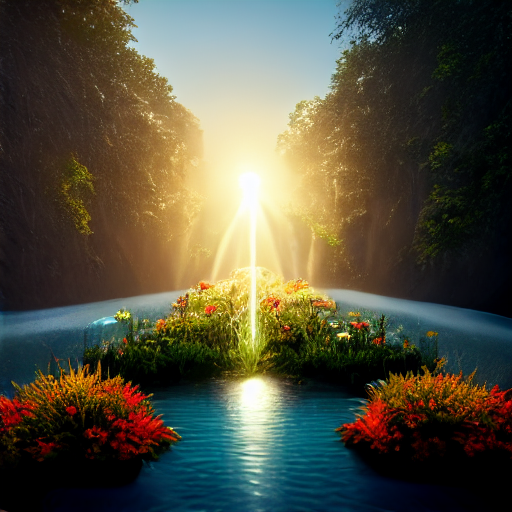} &
    \includegraphics[width=0.25\linewidth]{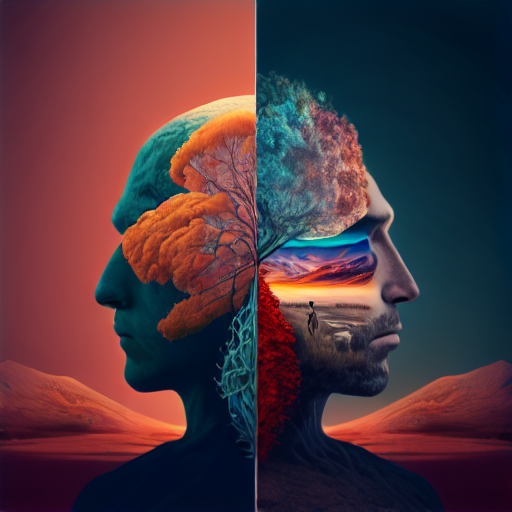} &
    \includegraphics[width=0.25\linewidth]{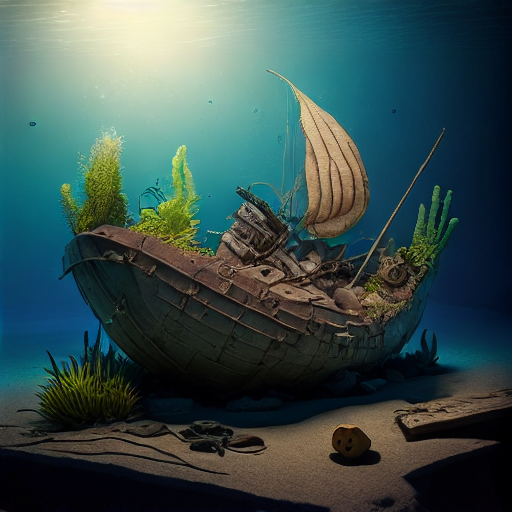} &
    \includegraphics[width=0.25\linewidth]{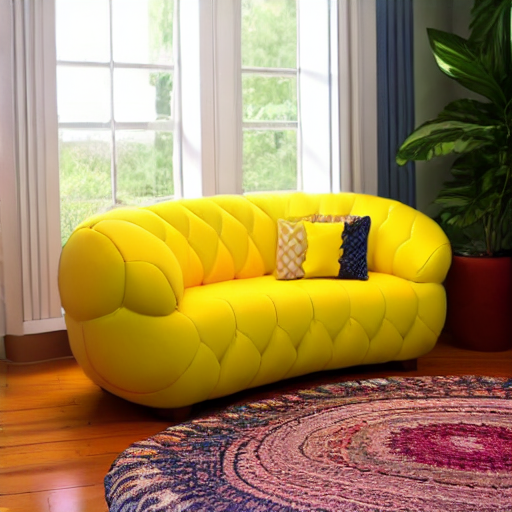}
    \\
    \includegraphics[width=0.25\linewidth]{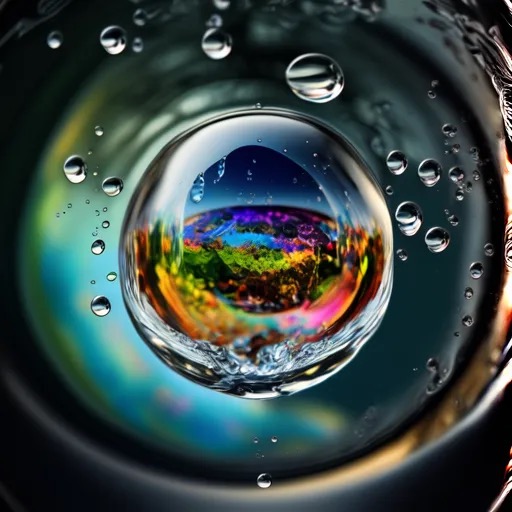} &
    \includegraphics[width=0.25\linewidth]{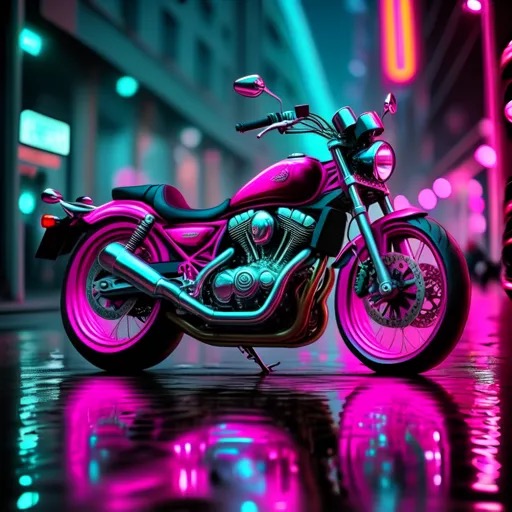} &
    \includegraphics[width=0.25\linewidth]{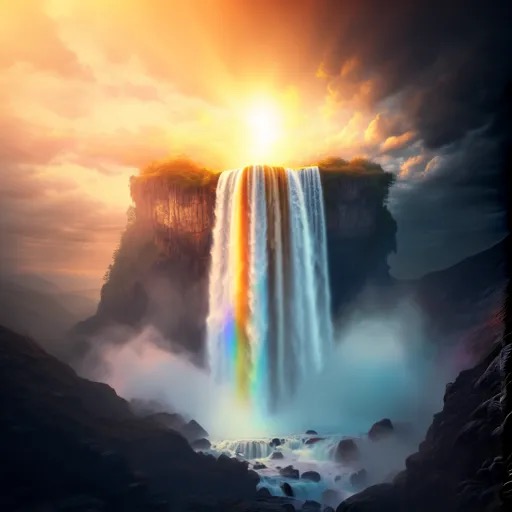} &
    \includegraphics[width=0.25\linewidth]{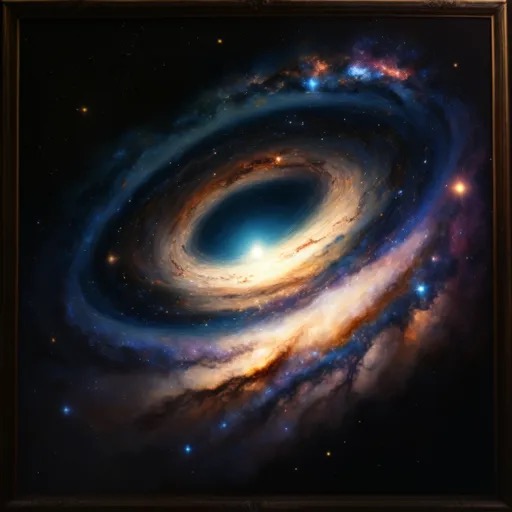}
    \\ 
    \includegraphics[width=0.25\linewidth]{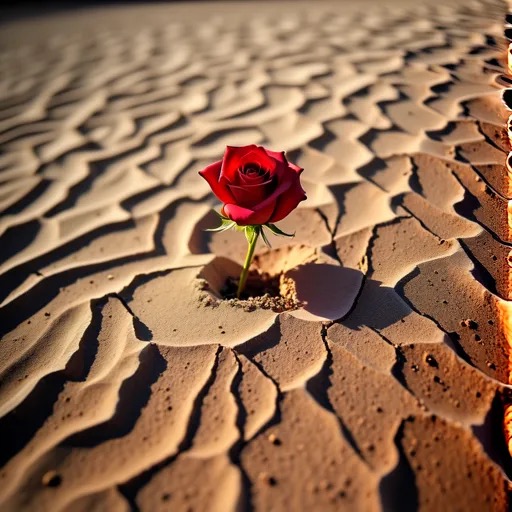} &
    \includegraphics[width=0.25\linewidth]{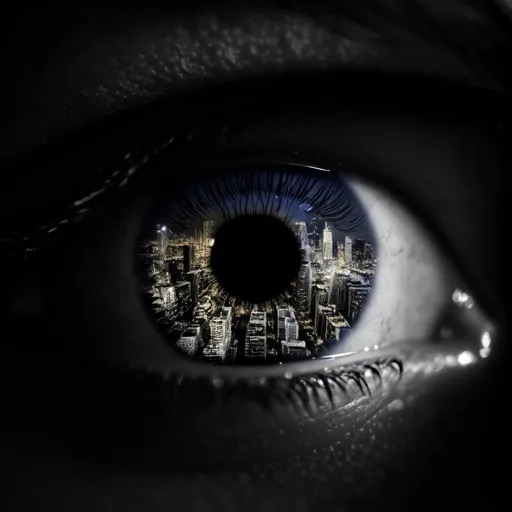} &
    \includegraphics[width=0.25\linewidth]{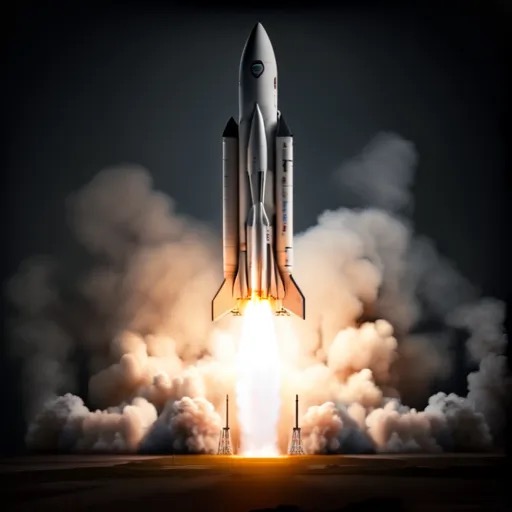} &
    \includegraphics[width=0.25\linewidth]{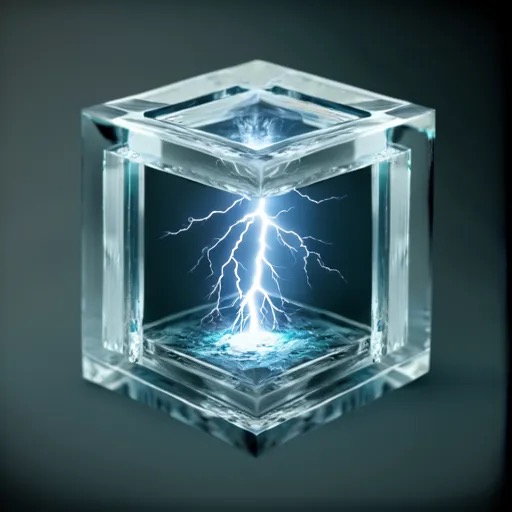}
    \\ 
    \includegraphics[width=0.25\linewidth]{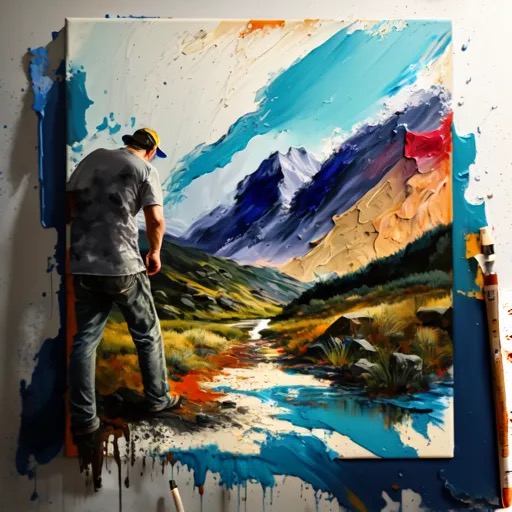} &
    \includegraphics[width=0.25\linewidth]{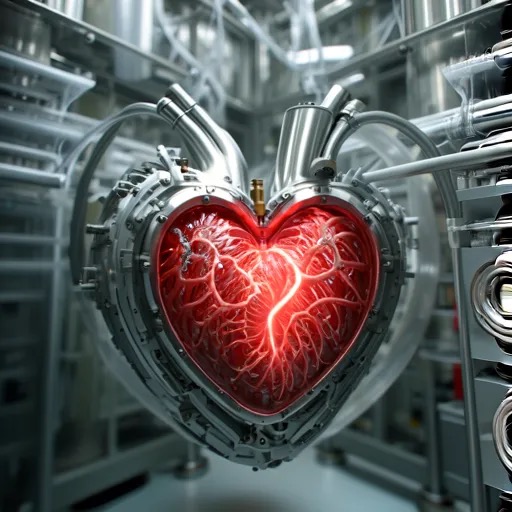} &
    \includegraphics[width=0.25\linewidth]{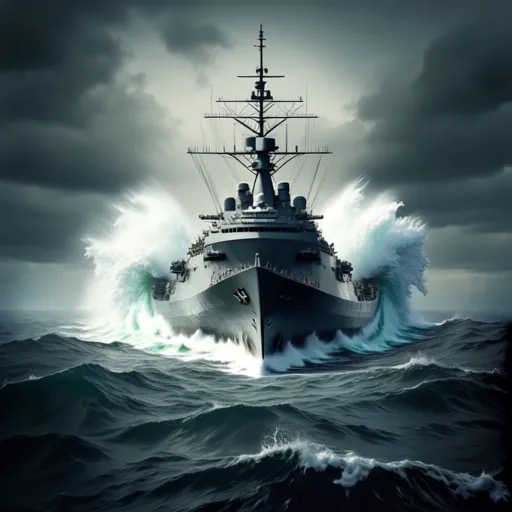} &
    \includegraphics[width=0.25\linewidth]{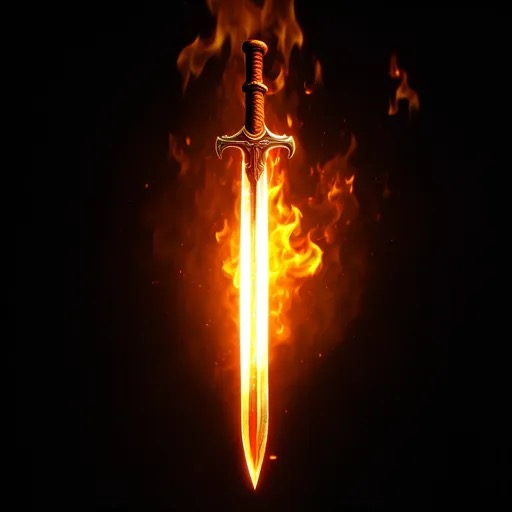}
  \end{tabular}
  \vspace{-0.2cm}
  \caption{\textbf{Qualitative text-to-image results.}}
  \label{fig:supp:t2i-grid}
\end{figure}

\begin{figure*}[tb]
    \centering
    \begin{minted}[
        fontsize=\scriptsize,
        linenos,
        breaklines,
        autogobble
    ]{diff}
 # Original file: https://github.com/willisma/SiT/blob/main/models.py
 
 def modulate(x, shift, scale):
-    return x * (1 + scale.unsqueeze(1)) + shift.unsqueeze(1)
+    return x * (1 + scale) + shift
 
 ...
 
 class SiTBlock(nn.Module):
     ...
     
     def forward(self, x, c):
-        shift_msa, scale_msa, gate_msa, shift_mlp, scale_mlp, gate_mlp = self.adaLN_modulation(c).chunk(6, dim=1)
-        x = x + gate_msa.unsqueeze(1) * self.attn(modulate(self.norm1(x), shift_msa, scale_msa))
-        x = x + gate_mlp.unsqueeze(1) * self.mlp(modulate(self.norm2(x), shift_mlp, scale_mlp))
+        shift_msa, scale_msa, gate_msa, shift_mlp, scale_mlp, gate_mlp = self.adaLN_modulation(c).chunk(6, dim=-1)
+        x = x + gate_msa * self.attn(modulate(self.norm1(x), shift_msa, scale_msa))
+        x = x + gate_mlp * self.mlp(modulate(self.norm2(x), shift_mlp, scale_mlp))
         return x
 ...

 class FinalLayer(nn.Module):
     ...
     
     def forward(self, x, c):
-        shift, scale = self.adaLN_modulation(c).chunk(2, dim=1)
+        shift, scale = self.adaLN_modulation(c).chunk(2, dim=-1)
         x = modulate(self.norm_final(x), shift, scale)
         x = self.linear(x)
         return x

 ...

 class SiT(nn.Module):
     def __init__(self, ...):
         ...
-        self.out_channels = in_channels
+        self.out_channels = in_channels + 1
     
     def forward(self, x, t, y):
         """
         t: (b, n) with n = number of tokens
         """
         x = self.x_embedder(x) + self.pos_embed
         
-        t = self.t_embedder(t)
+        # patch-level t's
+        t = t[..., None]                        # (b, n) -> (b, n, 1)
+        t = self.t_embedder(t)                  # (b, 1, n, d)
+        t = t.squeeze(1)                        # (b, n, d): one embedding per patch

-        y = self.y_embedder(y, self.training)
+        # broadcast y-embedding
+        y = self.y_embedder(y, self.training)   # (b, c)
+        y = y.unsqueeze(1)                      # (b, 1, c)

         c = t + y
         for block in self.blocks:
             x = block(x, c)
         x = self.final_layer(x, c)
         x = self.unpatchify(x)
         
-        return x
+        # split uncertainty
+        logvar_theta = x[:, -1:, :, :]          # (b, 1, h, w)
+        x = x[:, :-1, :, :]                     # (b, c, h, w)
+        return x, logvar_theta
    \end{minted}
    \caption{\textbf{Code changes} to adapt the original PyTorch SiT/DiT architecture~\cite{ma2024sit, peebles2023scalable-dit} to allow different noise scales per patch. Original file sourced from \url{https://github.com/willisma/SiT/blob/main/models.py}.}
    \label{fig:supp:code-sit}
\end{figure*}

\begin{figure*}[t]
\newcommand{\subfigy}[2]{%
  \begin{subfigure}{\textwidth}
    \footnotesize
    \centering
    \captionsetup{labelformat=empty}
    \includegraphics[width=1\linewidth]{fig/supp/uncurated/repa-pf-xl_cls#1.jpeg}
    \caption{#2}
  \end{subfigure}
}
\begin{minipage}{0.48\linewidth}
    \subfigy{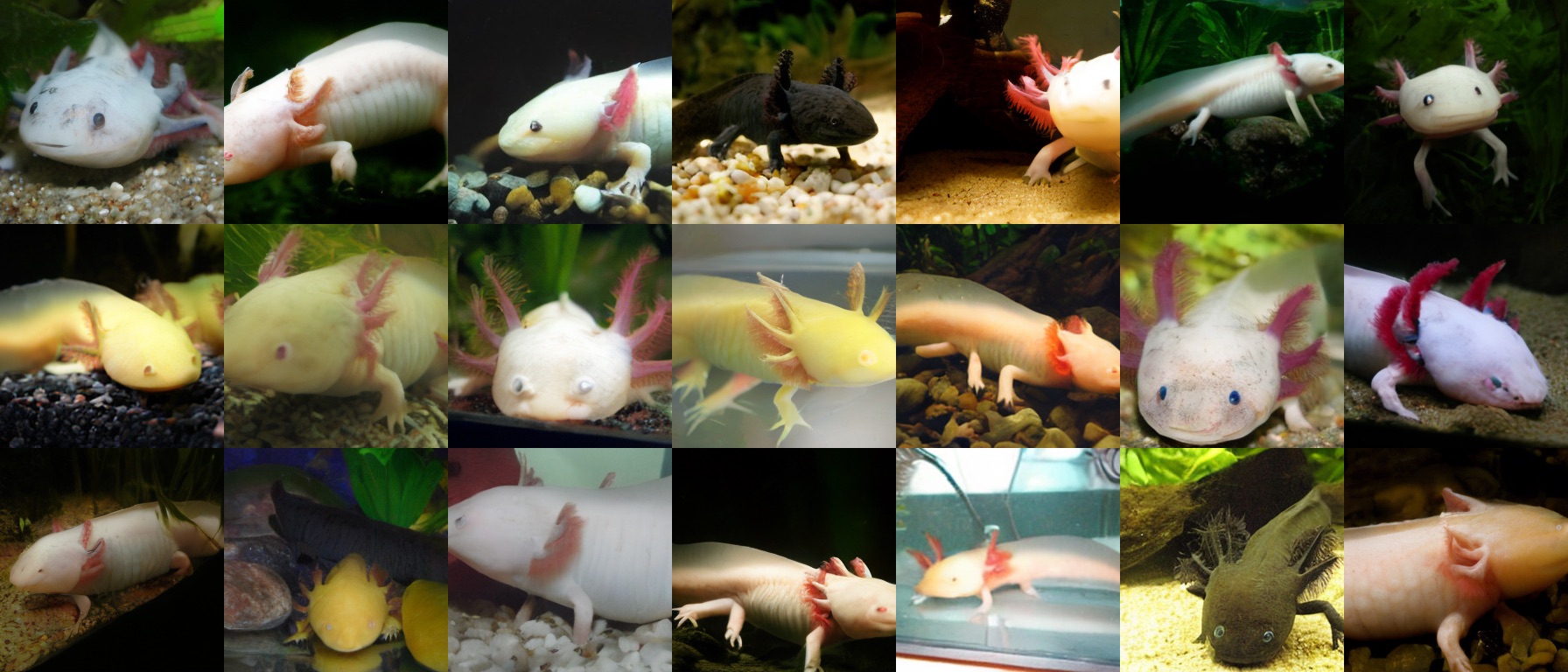}{29: Axolotl}
    \subfigy{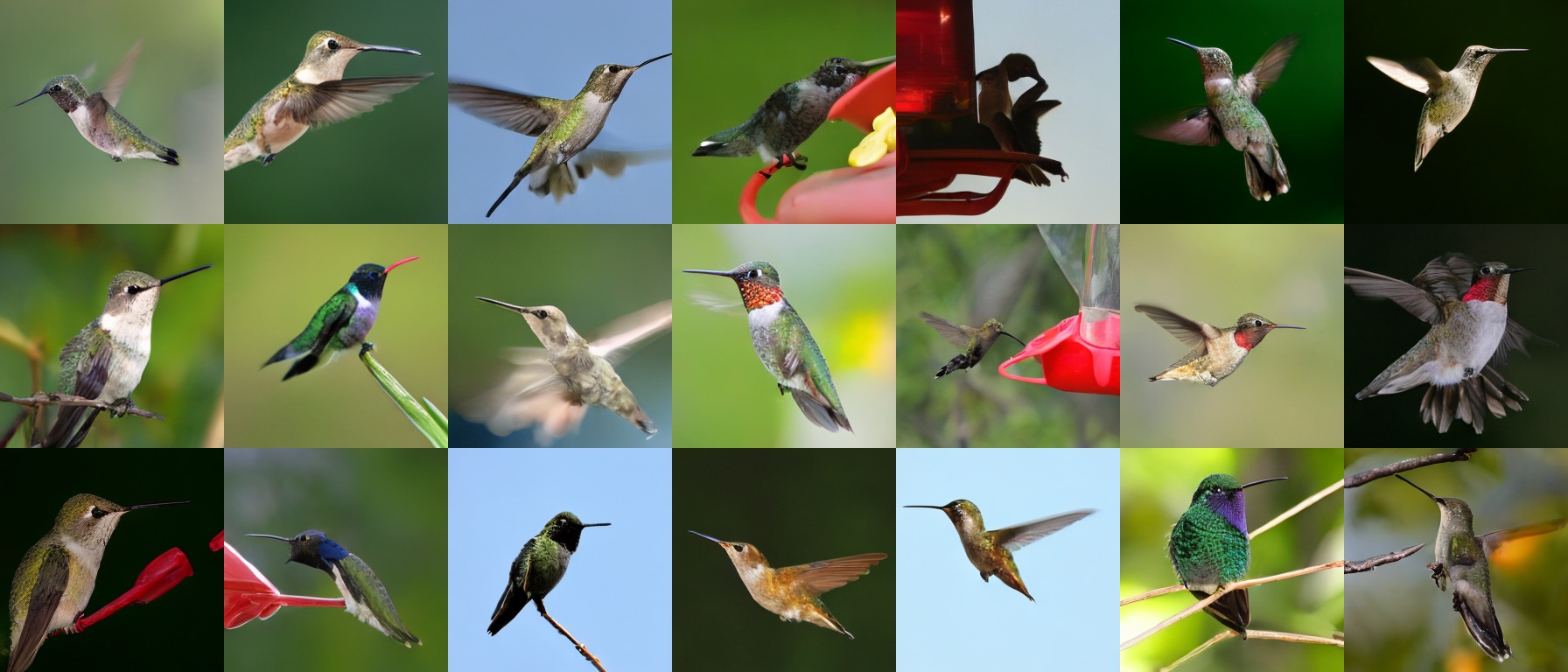}{94: Hummingbird}
    \subfigy{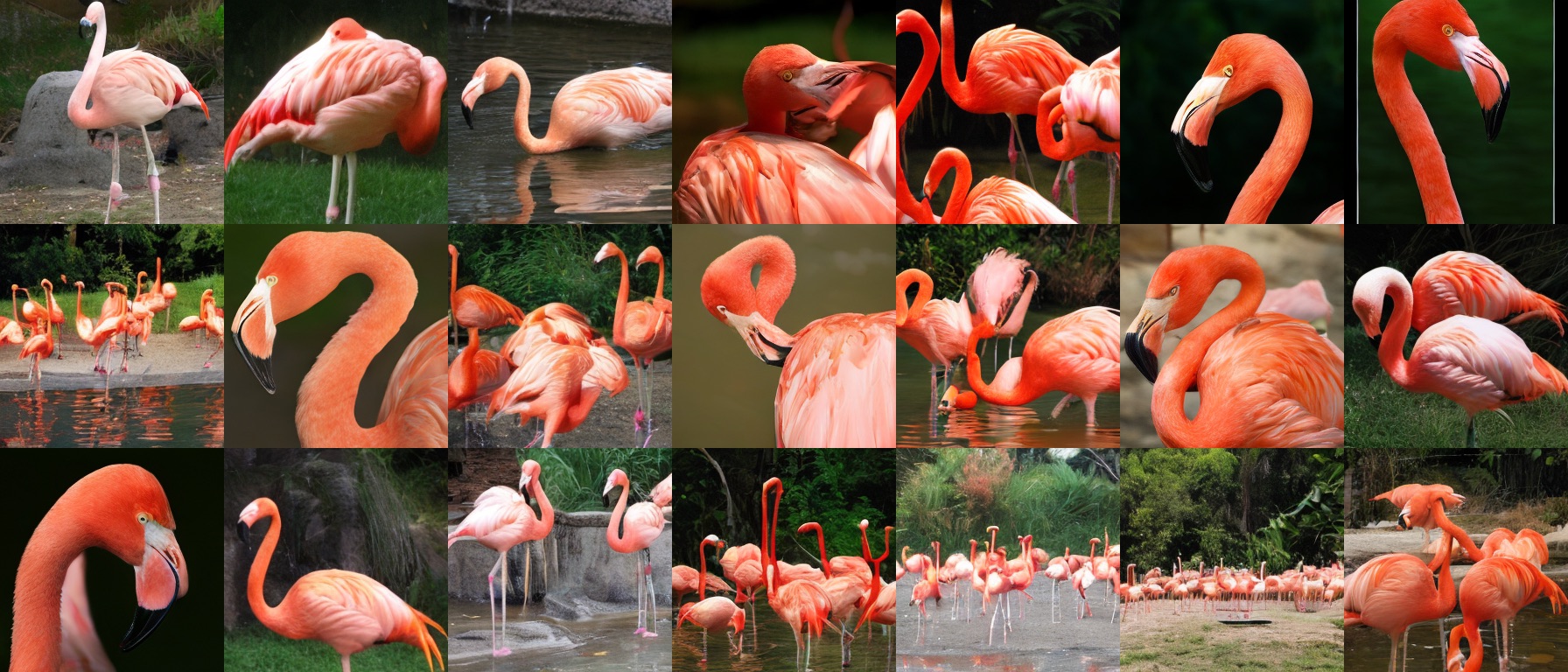}{130: Flamingo}
    \subfigy{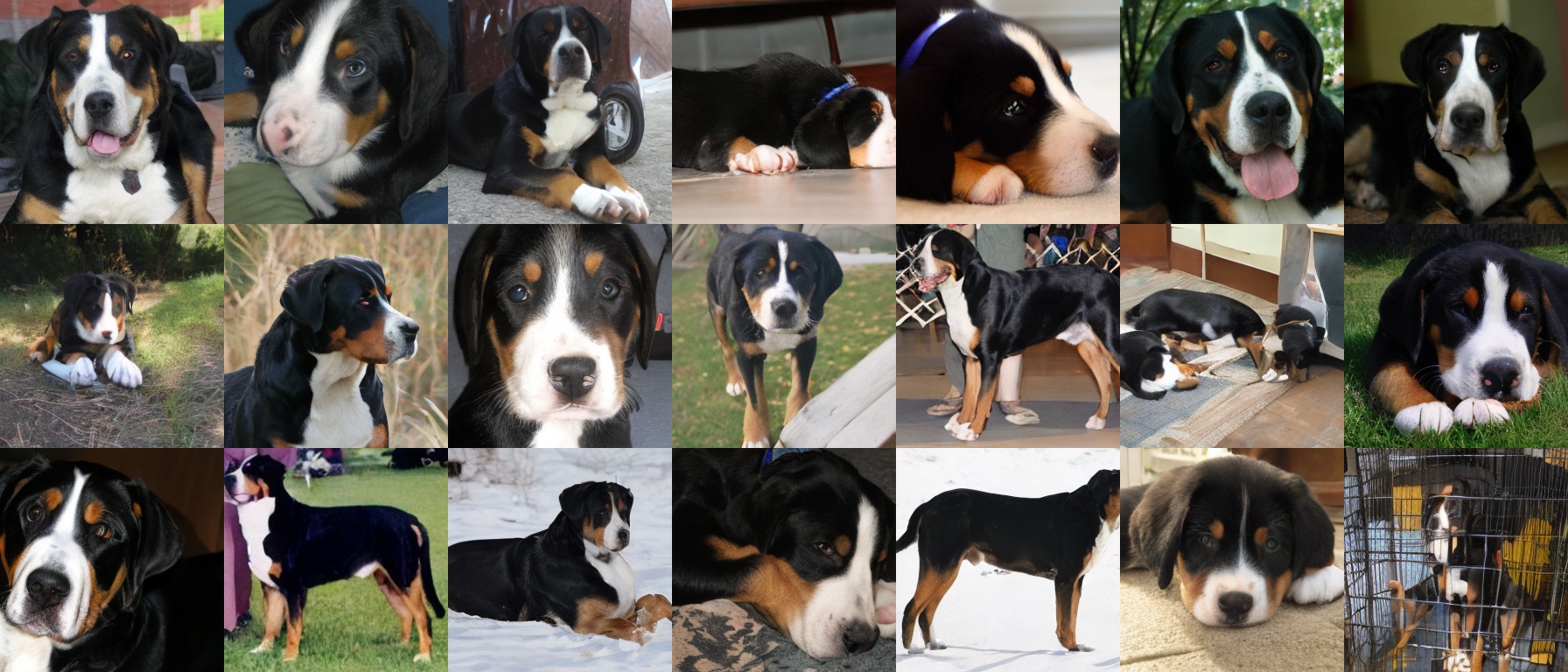}{238: Greater Swiss Mountain Dog}
    \subfigy{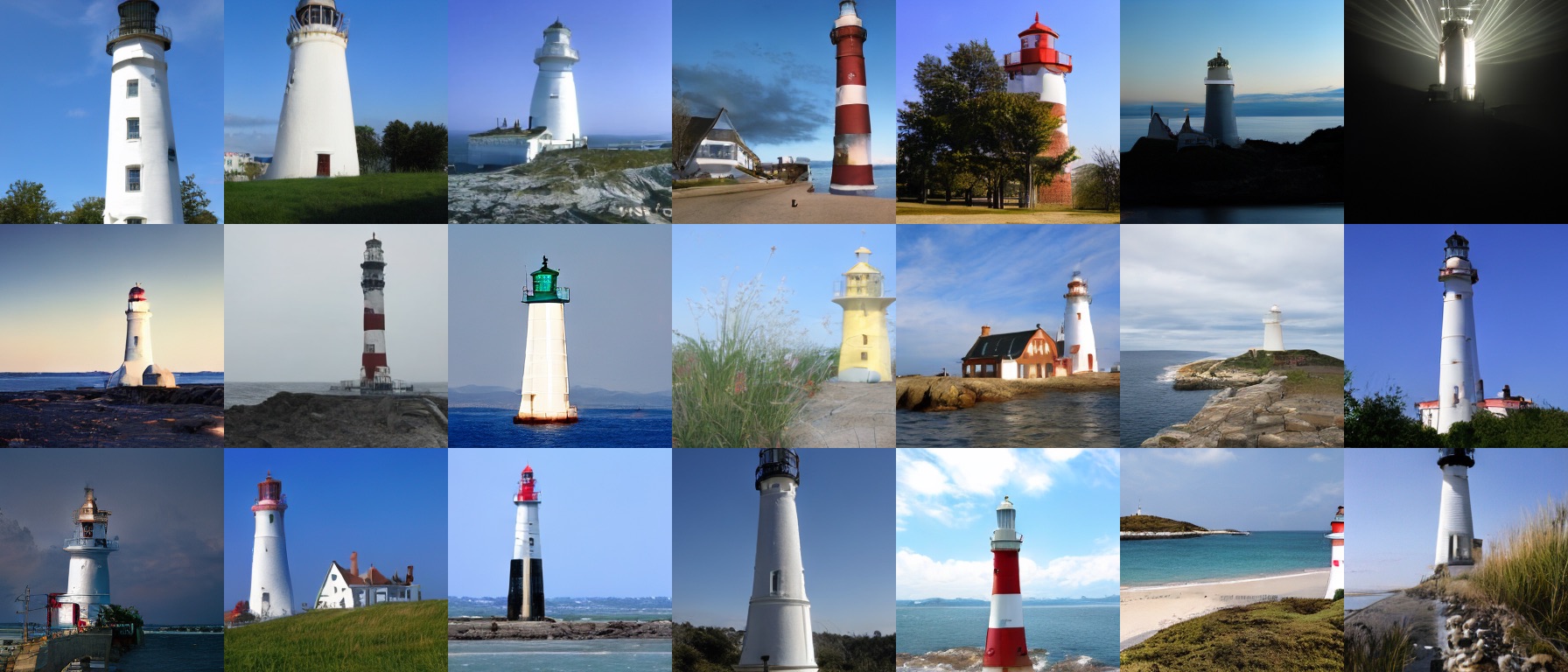}{437: Beacon}
\end{minipage}
\hfill
\begin{minipage}{0.48\linewidth}
    \subfigy{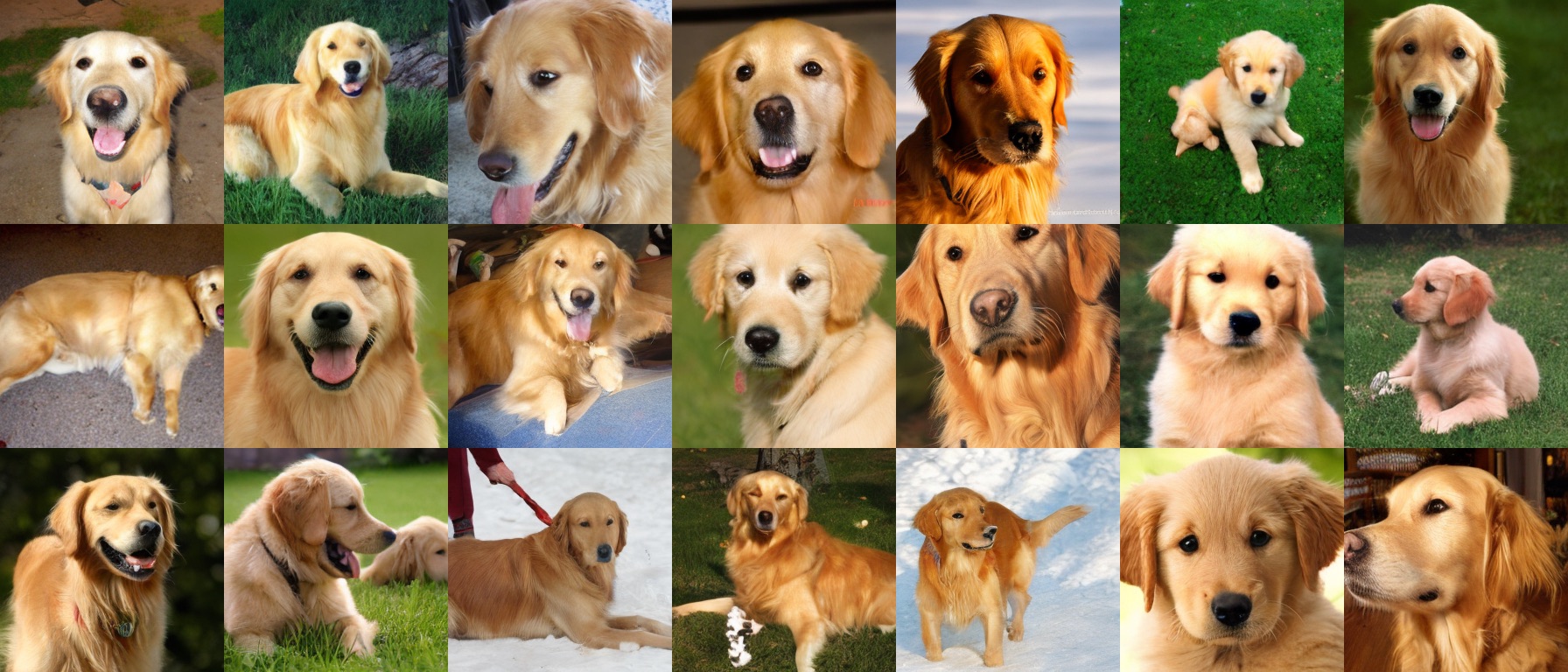}{207: Golden Retriever}
    \subfigy{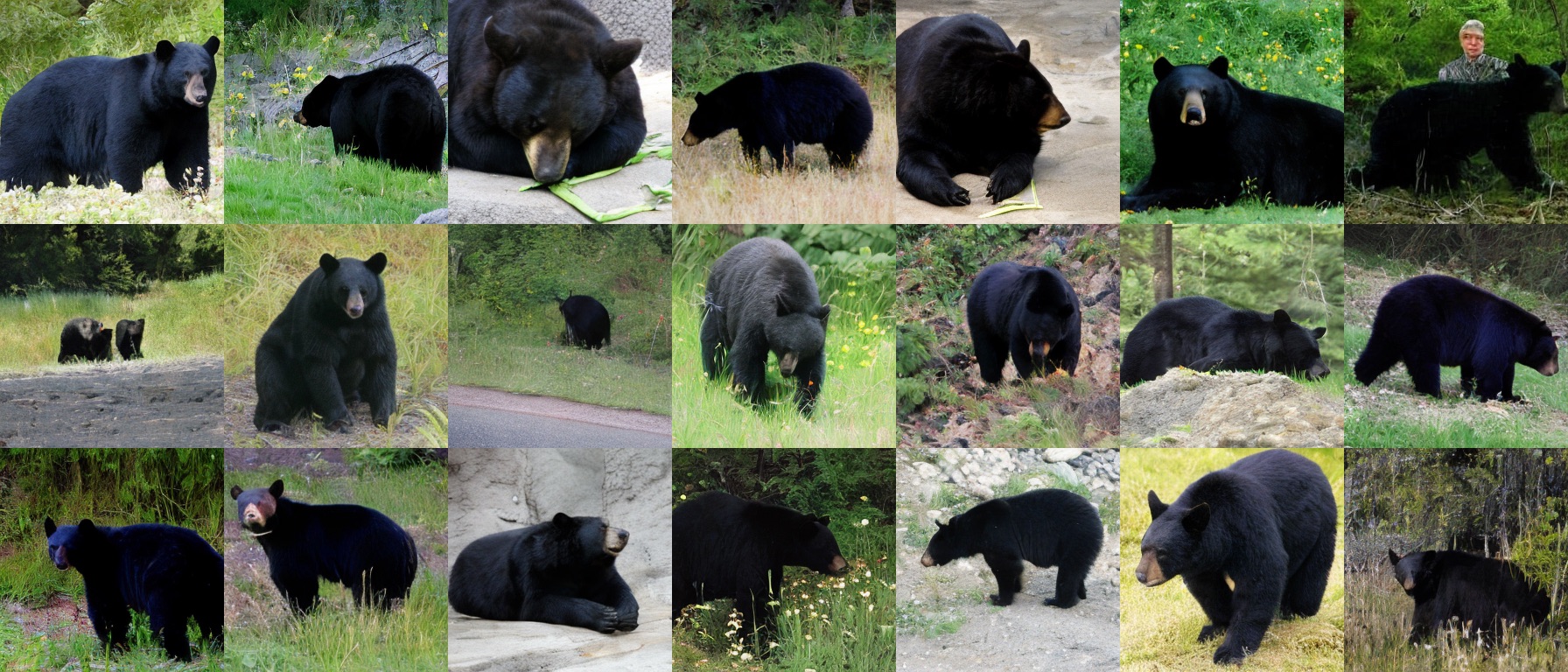}{295: American Black Bear}
    \subfigy{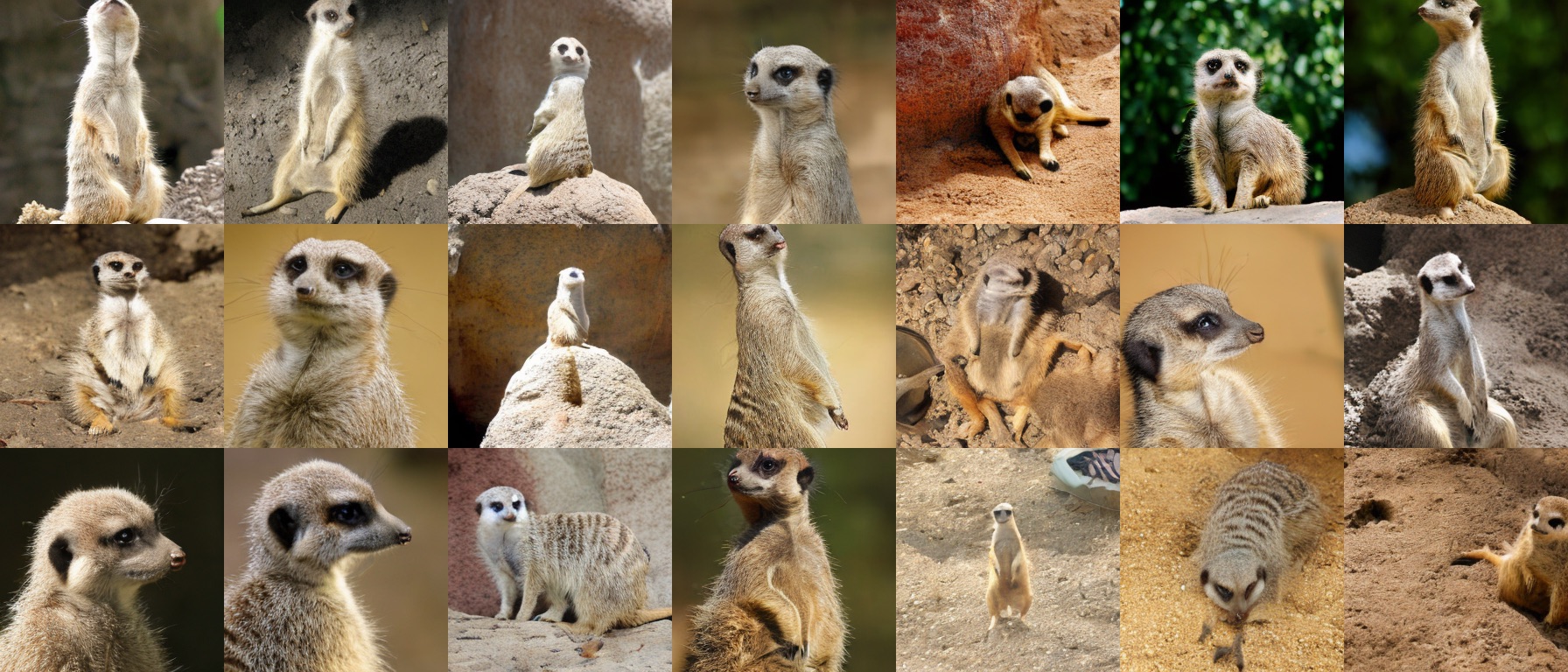}{299: Meerkat}
    \subfigy{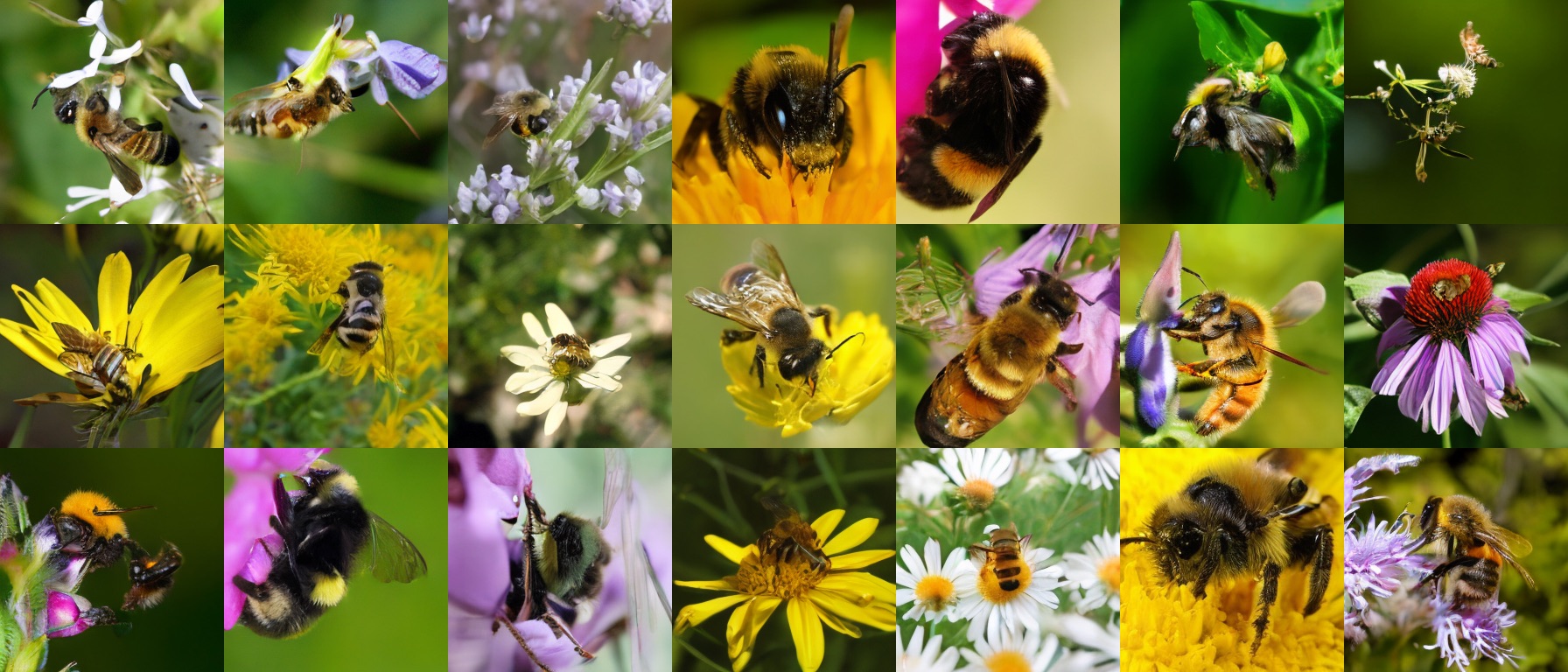}{309: Bee}
    \subfigy{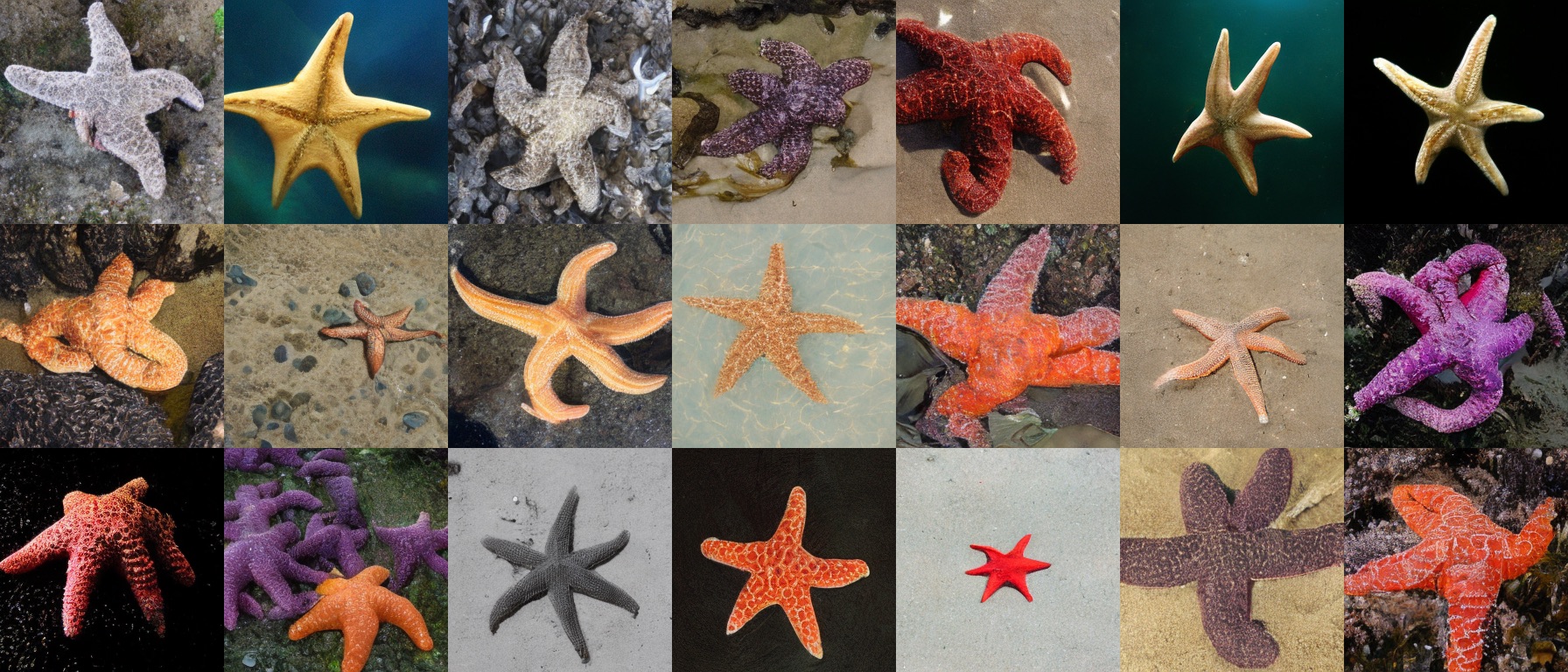}{327: Starfish}
\end{minipage}
\vspace{-0.6cm}
\caption{\textbf{Uncurated Imagenet $256 \times 256$ samples} from our REPA-PF-XL model. We use $100$ NFE and a CFG value of $2.5$.}
\label{fig:supp:qualitative-imagenet}
\end{figure*}

\end{document}